\pdfoutput=1

\documentclass[11pt,table]{article}

\usepackage[]{EMNLP2023}

\usepackage{times}
\usepackage{latexsym}
\usepackage{graphicx}
\usepackage{amsmath}

\usepackage{booktabs}
\usepackage{multirow}
\usepackage{subcaption}
\usepackage{arydshln}
\usepackage{soul}
\usepackage{algorithm}
\usepackage{algpseudocode}
\usepackage{rotating}
\usepackage{graphicx}
\usepackage{booktabs}
\definecolor{lightblue}{rgb}{0.68, 0.85, 0.9}
\usepackage{comment}

\usepackage[T1]{fontenc}

\usepackage[utf8]{inputenc}

\usepackage{microtype}

\usepackage{inconsolata}

\definecolor{todo}{rgb}{1,0.5,0}
\usepackage{enumitem}

\title{
Are Large Language Models Good Classifiers?  \\ A Study on Edit Intent Classification in Scientific Document Revisions
}

\author{Qian Ruan, Ilia Kuznetsov, Iryna Gurevych  \\
        Ubiquitous Knowledge Processing Lab (UKP Lab)\\
        Department of Computer Science and Hessian Center for AI (hessian.AI)\\
        Technical University of Darmstadt \\
  \texttt{www.ukp.tu-darmstadt.de}}

\begin{document}
\captionsetup[table]{skip=5pt}
\captionsetup[figure]{skip=5pt}
\maketitle
\begin{abstract}
Classification is a core NLP task architecture with many potential applications. While large language models (LLMs) have brought substantial advancements in text generation, their potential for enhancing classification tasks remains underexplored. 
To address this gap, we propose a framework for thoroughly investigating fine-tuning LLMs for classification, including both generation- and encoding-based approaches.
We instantiate this framework in edit intent classification (EIC), a challenging and underexplored classification task.
Our extensive experiments and systematic comparisons with various training approaches and a representative selection of LLMs yield new insights into their application for EIC. We investigate the generalizability of these findings on five further classification tasks.
To demonstrate the proposed methods and address the data shortage for empirical edit analysis,
we use our best-performing EIC model to create \textit{Re3-Sci2.0}, a new large-scale dataset of 1,780 scientific document revisions with over 94k labeled edits. The quality of the dataset is assessed through human evaluation.
The new dataset enables an in-depth empirical study of human editing behavior in academic writing.
We make our experimental framework\footnote{\url{https://github.com/UKPLab/llm_classifier}}, models and data\footnote{\url{https://tudatalib.ulb.tu-darmstadt.de/handle/tudatalib/4355}} publicly available.

\end{abstract}
\section{Introduction}
\label{sec:intro}
Generative large language models (LLMs)  have demonstrated substantial advancements in text generation tasks \cite{qa_llm2, summ_llm, summ_llm2}. 
However, their potential for enhancing classification tasks, a significant subset of NLP applications, remains underexplored. 
The predominant strategy for applying LLMs to classification tasks is to cast them as generation tasks, followed by instruction tuning \cite{cls_llm7, cls_llm, cls_llm3, cls_llm6, cls_llm8}, supervised fine-tuning \cite{cls_llm5}, and active learning \cite{cls_llm2}, all of which aim to generate label strings within the output tokens.
\begin{figure}[t]
  \centering
    \includegraphics[width=0.55\textwidth]{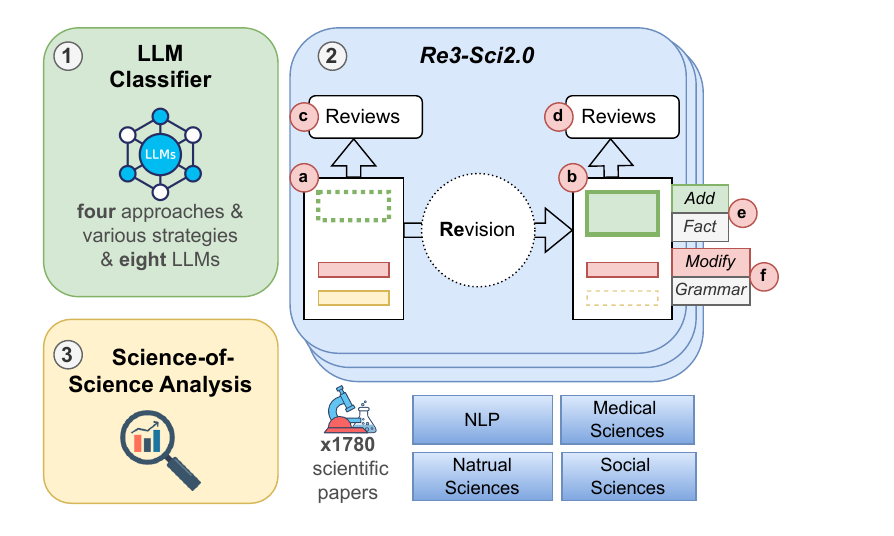}
    \caption{In this work, we (1). present a general framework to explore the classification capabilities of LLMs, conducting extensive experiments and systematic comparisons on the EIC task;
    (2). use the best model to create the \textit{Re3-Sci2.0} dataset, which comprises 1,780 scientific document revisions (a-b), associated reviews (c, d), and 94,482 edits annotated with action and intent labels (e, f), spanning various scholarly domains;
    (3). provide a first in-depth empirical analysis of human editing behavior using this new dataset.}
    \label{fig:overview}
\end{figure}
Recent studies \cite{nvembed, Linq, sfr} have shown the superiority of LLMs as embedding models on the MTEB benchmark \cite{mteb}.
However, there is a lack of a holistic framework for a systematic study of the classification capabilities of LLMs in end-to-end fine-tuning paradigms. 
Yet, such a framework is important as it extends beyond the current use of LLMs as generative or embedding models for classification, opens new opportunities for a wide range of real-world tasks, and reveals 
novel potential for advanced LLM training and utilization.

To instantiate the framework, we seek a \textbf{complex}, \textbf{challenging}, and \textbf{underexplored} task that is \textbf{crucial} for addressing unresolved real-world applications.
Edit intent classification (EIC) is such a complex task, aiming to identify the purpose of textual changes, necessitating a deep understanding of the fine-grained differences between paired inputs.
Previous works have provided small human-annotated datasets and demonstrated the crucial role of the intent labels in studying domain-specific human editing behavior \cite{argrewrite, wikieditintent-yang, argrewrite2, re3}. 
However, due to the high cost of human annotation, existing datasets are limited in size. There is a lack of effective NLP automation and extensive labeled datasets to facilitate larger-scale revision analysis.
From the modeling perspective, previous studies have primarily explored EIC using basic feature engineering \cite{argrewrite, wikieditintent-yang, argrewrite2}, fine-tuning small pre-trained language models (PLMs) \cite{IteraTeR, arxivedits}, or instruction tuning with LLMs \cite{re3}. Advanced methodologies involving fine-tuning LLMs remain unexplored. 
The suboptimal results of previous works (Table \ref{tab:related_work}) further highlight the task's inherent difficulty and the necessity for advancements in NLP.\footnote{Note that direct performance comparison in Table \ref{tab:related_work} is not possible due to different datasets, label sets and data sizes, but they illustrate the inherent difficulty of EIC despite data variations.}

To close the gap, we introduce a general framework to explore the use of LLMs for classification, featuring one generation-based and three encoding-based fine-tuning approaches (§\ref{sec:method}).
We instantiate the framework in EIC, conduct extensive experiments and provide novel insights from systematic comparisons of the four approaches, eight LLMs, and various training strategies. 
Our findings reveal that LLMs fine-tuned with encoding-based approaches demonstrate superior classification capabilities for EIC, achieving state-of-the-art (SOTA) performance.  To demonstrate the versatility of our framework, we apply it to five further classification tasks and investigate the generalizability of our insights (§\ref{sec:results}). 
To illustrate the application of the proposed methods for EIC and address the lack of data for extensive edit analysis, we use our best-performing models to create \textit{Re3-Sci2.0}, a large-scale dataset with 1,780 scientific document revisions and 94,482 labeled edits across various research domains (§\ref{sec:auto_anno}). 
This dataset enables the first in-depth science-of-science \cite{scisci} analysis of scientific revision success and human editing behavior across research domains (§\ref{sec:meta_analysis}).
Our work thus makes four key \textbf{contributions}: 
\begin{itemize}[itemsep=0pt, parsep=0pt]
\item A general framework for fine-tuning LLMs for classification tasks, with four approaches and various training strategies. 
\item Extensive experiments on EIC, and systematic comparisons of different approaches, training strategies, base PLMs and LLMs,\footnote{While current LLM terminology requires further precision, as discussed in \citet{llms_terms_claims}, we use the terms "LLMs" and "PLMs" for readability. "LLMs" refers to latest-generation large-scale language models, such as Mistral-Instruct, LLaMA, and GPT-4, which cannot be trained or fully fine-tuned on one or two modern GPUs. In contrast, "PLMs" denotes earlier smaller pre-trained language models, such as T5, RoBERTa, and other BERT variants, which can be trained and fully fine-tuned using one or two GPUs. Details on the language models are provided in §\ref{subsec:lms}} supplemented by evaluation on five further classification tasks.
\item A large dataset of 1,780 scientific document revisions with 94,482 edits, annotated by our best EIC model, which achieves a macro average F1 score of 84.3.
\item A first in-depth science-of-science analysis of scientific revision success and human editing behavior across various scholarly domains.
\end{itemize}
Our work paves the path towards systematically investigating the use of LLMs for classification tasks. 
Our experiments yield substantial results in the challenging EIC task. The resulting large-scale dataset facilitates empirical analysis of human editing behavior in academic publishing and beyond.

\section{Related Work}
\label{sec:related_work}

\begin{table}[h]
\tabcolsep=0.06cm
\fontsize{8}{8}
\selectfont
\tabcolsep=0.09cm
\renewcommand{\arraystretch}{1.2}
\begin{tabular}{llllllll} \toprule
      &\#label &\#train&\#test & acc.
       &method\\ \hline
       \citet{argrewrite} &8&1,757&10CV&58.8*&FE\\
        \citet{wikieditintent-yang} &13&5,777&10CV&59.7*&FE \\ 
        \citet{argrewrite2} &9&3,238&5CV&68&FE\\
       \citet{IteraTeR} &5&3,254&364&49.4*&PLM \\ 
       \citet{arxivedits}&4&600&200&84.4&PLMs \\
       \citet{arxivedits}&9&600&200&79.3&PLMs \\
      \citet{re3} & 5 &2,234&8,936&70&LLM (inst) \\
        Ours & 5 &7,478&2,312 &85.6 & PLMs \& LLMs \\ \bottomrule
       \end{tabular}
       \caption{Comparison of related works on EIC, including counts of unique intent labels, training and test samples, best accuracy (or *macro average F1 scores), and explored methods. nCV: n-fold cross-validation. FE: feature engineering. 
       } 
       \label{tab:related_work}
\end{table}
\noindent\textbf{Edit Intent Classification.} 
Identifying the underlying intent of textual edits is a challenging yet underexplored task, with only a few studies contributing taxonomies, datasets and methodologies. 

While existing works \cite{argrewrite, wikieditintent-yang, argrewrite2, IteraTeR, arxivedits, re3} demonstrate the critical role of intent labels in understanding human editing, they also highlight the challenges of manually labeling edit intent, which requires expert annotators, specialized annotation tools, and extensive training \cite{arxivedits, re3}. The high costs and efforts of manual labeling limit the size of available datasets (Table \ref{tab:related_work}), restrict large-scale studies of human editing behavior and motivate the need for effective NLP automation in EIC and the creation of larger labeled datasets. 

From the modeling perspective, several works \cite{argrewrite, wikieditintent-yang, argrewrite2} have primarily investigated automatic EIC using various feature engineering techniques and employed basic classifiers such as SVM \cite{svm}, MULAN \cite{mulan}, and XGBoost \cite{XGBoost}.  
Other studies \cite{IteraTeR, arxivedits} explored fine-tuning PLMs such as RoBERTa \cite{roberta}, T5 \cite{t5}, and PURE \cite{PURE}.
\citet{re3} is the first application of LLMs for EIC. However, it is limited to using Llama2-70B \cite{llama2} with instruction tuning, without any fine-tuning.

As outlined in Table \ref{tab:related_work}, our work is the first to systematically compare different fine-tuning approaches for a broad set of PLMs and LLMs using various training strategies for EIC, achieving substantial progress in this challenging NLP task (§\ref{sec:method}). To address the shortage of revision datasets, we use our most efficient and high-performing EIC model to create the new, large-scale Re3-Sci2.0 dataset. We provide a comprehensive overview of the entire pipeline — model selection (§\ref{sec:results}), annotation (§\ref{sec:auto_anno}), and revision analysis (§\ref{sec:meta_analysis}) — to ensure a complete and reproducible process for generating high-quality, large-scale automatically labeled revision datasets with LLMs.

\begin{figure*}[ht]
  \centering
  \includegraphics[width=0.76\textwidth]{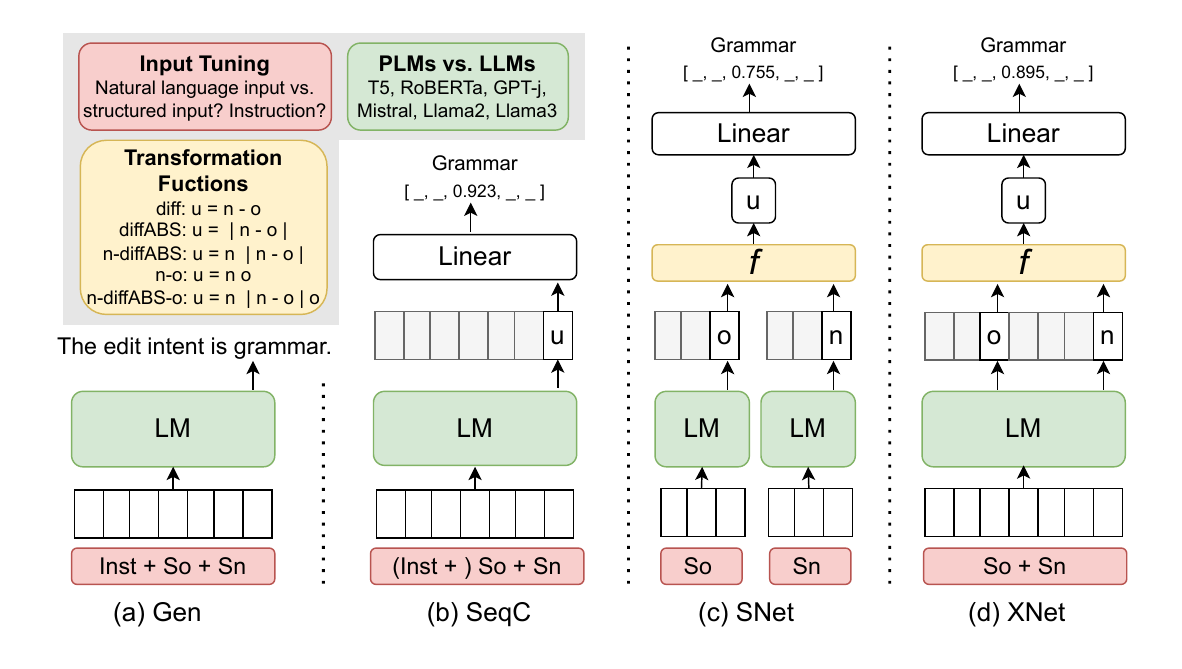}
  \caption{Proposed approaches with a systematic investigation of the key components: input types (red), language models (green), and transformation functions (yellow). See §\ref{sec:method} and §\ref{sec:results} for details.}
  \label{fig:approaches}
\end{figure*}

\noindent\textbf{LLMs for Classification.} 
Previous studies have utilized LLMs for classification, primarily aiming to generate label strings within the output tokens through instruction tuning \cite{cls_llm7, cls_llm, cls_llm3, cls_llm6, cls_llm8}.
Few studies have enhanced LLMs to generate label text through supervised fine-tuning \cite{cls_llm5} and active learning \cite{cls_llm2}.
Additionally, recent studies \cite{nvembed,Linq,sfr} have demonstrated the superiority of LLMs as embedding models on MTEB\footnote{\url{https://huggingface.co/blog/mteb}} \cite{mteb}, an extensive text embedding benchmark where embeddings are processed by additional classifiers.
However, there is a lack of a holistic framework for systematically investigating the encoding capabilities of LLMs in end-to-end fine-tuning paradigms.
We are the first to address the gap by proposing encoding-based methodologies that extensively investigate and fine-tune LLMs as supervised classification models, systematically comparing these methodologies with the generation-based approach within a unified framework.
While our work focuses on the challenging and crucial  
EIC task (§\ref{sec:intro}), our framework is applicable to a wide range of classification tasks, as demonstrated by our experiments with additional tasks in §\ref{subsec:gen_eval}.

\section{Framework} 
\label{sec:method} 
We investigate four distinct approaches to fine-tune LLMs for classification (§\ref{subsec:approaches}),  use various training strategies including three input types (§\ref{subsec:input_tuning}) and five transformation functions (§\ref{subsec:trans_func}), systematically comparing different language models (§\ref{subsec:lms}).

\subsection{Approaches}
\label{subsec:approaches}
We illustrate the proposed approaches to text classification using the EIC task. We formulate it as a multi-class pair classification task involving a sentence edit pair $e(S_o, S_n)$, where $S_o$ represents the original sentence and $S_n$ denotes the new sentence after the edit. In cases of sentence additions or deletions, only the single added/deleted sentence ($S_n$/$S_o$) is provided, while the corresponding pair sentence remains empty. The objective is to predict an edit intent label $l$ from a set of $k$ possible labels $L$. As illustrated in Figure \ref{fig:approaches}, 
\begin{itemize}[itemsep=1.5pt, parsep=-0.5pt]
\item \textbf{Approach \textit{Gen}} addresses the task as a text generation task, aiming to produce the label string within the output tokens from input text that includes the task instruction, the old sentence $S_o$, and the new sentence $S_n$. 
\item \textbf{Approach \textit{SeqC}} treats the task as a sequence classification task using LLMs equipped with a linear classification layer on top. It utilizes the last hidden states of the last token ($u$) as the input embedding for classification. The linear layer transforms $u$ of the model size $d$ into a $k$-dimensional logit vector, where the maximum value indicates the predicted label.
\item \textbf{Approach \textit{SNet}} employs a Siamese architecture akin to SBERT \cite{sbert} for sequence classification. It processes the two sentences independently through twin Siamese LLMs, producing $o$ and $n$ (representing the last token of each), for the old and new sentences respectively. A transformation function $f$ (§\ref{subsec:trans_func}) converts these into a single representation $u$ for classification.
\item \textbf{Approach \textit{XNet}} employs a cross network to process both sentences simultaneously through a single LLM, extracting the last-token embeddings $o$ and $n$ for the old and new sentences respectively. They are then transformed into a single representation $u$ by a function $f$ for classification.
\end{itemize}

\subsection{Input Tuning}
\label{subsec:input_tuning}
The input text, indicated by red blocks in Figure \ref{fig:approaches}, comprises three components:  the task instruction (\textit{inst}), the original sentence $S_o$ and the new sentence $S_n$. The task instruction outlines the task's objective and specifies the possible labels. 
The input text is provided in two different formats: (1) \textit{natural input}, which includes only the content of the instruction and the sentences, and (2) \textit{structured input}, where the content is enclosed within specific structure tokens such as \textit{<instruction></instruction>}, \textit{<old></old>}, and \textit{<new></new>}.
In our experiments, we tune the presence of task instructions
and the input text formats to explore their effects (§\ref{sec:results}). Examples of input texts are displayed in Table \ref{tab:input_examples} in §\ref{sec:method_app}.

\subsection{Transformation Functions}
\label{subsec:trans_func}
In approaches \textit{SNet} and \textit{XNet}, the representations of the old and new sentences, $o$ and $n$, can be transformed into a single representation $u$ using five different transformation functions $f$: 
\begin{equation}
\label{eq:diff}
f_{diff}: u = n - o
\end{equation}
\begin{equation}
\label{eq:diffABS}
f_{diffABS}: u = | n - o |
\end{equation}
\begin{equation}
\label{eq:n-diffABS}
f_{n-diffABS}: u = n \oplus | n - o |
\end{equation}
\begin{equation}
\label{eq:n-o}
f_{n-o}: u = n \oplus o
\end{equation}
\begin{equation}
\label{eq:n-diffABS-o}
f_{n-diffABS-o}: u = n \oplus | n - o | \oplus o
\end{equation}
where $\oplus$ represents vector concatenation, - denotes vector subtraction, and | | indicates that absolute values are taken from the subtraction.  
The intuition is to incorporate the differences between sentence embeddings in the transformation process, as the EIC task relies on analyzing the variations between two versions of a text.
The five proposed transformation functions are systematically evaluated in our experiments (§\ref{sec:results}).

\begin{table*}[ht]
\fontsize{8}{8}
\selectfont
\centering
\renewcommand{\arraystretch}{1.2} 
\tabcolsep=0.12cm
\begin{subtable}[h]{0.999\linewidth}
\begin{tabular}[]{lllllllllllll}
\toprule
\multicolumn{13}{c}{\textbf{Baselines}} \\ \hline
& size & acc. & m. f1 &AIR && acc. & m. f1 &AIR \\ \cline{3-5}\cline{7-9}
Human &-&90.2&89.7&100\\ \cline{3-5}\cline{7-9}
&&\multicolumn{3}{c}{zero-shot} &
\multicolumn{4}{c}{ICT+CoT}\\ \cline{3-5}\cline{7-9}
GPT-4 &- & 45.5 &37 &99.9&&64.8&60.9&100\\
Llama2-70B \citeyearpar{re3} & 70B &-&-&-&& 70\textsuperscript{$\dagger$} & 69\textsuperscript{$\dagger$} &100\\
\hline
\multicolumn{13}{c}{\textbf{(a). \textit{Gen}}} \\ \hline
&&\multicolumn{3}{c}{\textbf{NFT Baselines}} &
\multicolumn{8}{c}{\textbf{Fine-tuned Models}}\\ \cline{7-13}
&&&&&&\multicolumn{3}{l}{\textcircled{\tiny 1} \textit{ inst + natural input}}
&&\multicolumn{3}{l}{\textcircled{\tiny 2} \textit{ inst + structured input}}\\
\cline{3-5} \cline{7-9} \cline{11-13} 
base LM & size & acc. & m. f1&AIR && acc. & m. f1 &AIR  && acc. & m. f1 &AIR\qquad\qquad\qquad\\ \cline{3-5} \cline{7-9} \cline{11-13}  
T5 & 220M&1.2&1.8&4.8&&\underline{79.9}&\underline{78.1}&100&&78.3 ($\downarrow$1.6)&78.0 
 ($\downarrow$0.1) &100\\
GPT-j & 6B& 12.6&11.2&68.9&&\underline{32.8}&\underline{21.0}&97.6&&21.2 ($\downarrow$11.6)
&15.4 ($\downarrow$5.6)&86.8 ($\downarrow$10.8)\\
Mistral-Instruct & 7B&28.0&24.0&99.9&&\underline{68.5}&\underline{63.4}  &100&& 62.8 ($\downarrow$5.7)
& 59.2 ($\downarrow$4.2)&100  \\
Llama2-7B & 7B &21.4 &12.2&78.2&&34.3  &24.7 &100&&\underline{60.4} ($\uparrow$26.1)
&\underline{47.6} ($\uparrow$22.9)&88.7 ($\downarrow$11.3)\\
Llama2-7B-Chat & 7B &12.1 &8.6 &85.2&&63.0  &49.2&100&&\underline{72.4} ($\uparrow$9.4)
&\underline{55.0} ($\uparrow$5.8)&88.5 ($\downarrow$11.5)\\
Llama2-13B & 13B &13.8&5.2&93.3 &&50.9 &39.5 &99.9&& \underline{73.4} ($\uparrow$22.5)
&\underline{67.6} ($\uparrow$28.1)&85.9 ($\downarrow$14.0)\\
Llama2-13B-Chat & 13B &0.5&1.9&2.0&&75.5&72.9&100&&\underline{83.6} ($\uparrow$8.1)
&\underline{82.8} ($\uparrow$9.9)&100\\ 
Llama3-8B & 8B &14.0&13.3&77.8&&79.4&79.1&95.4
&&\underline{83.3} ($\uparrow$3.9)&\underline{82.1} ($\uparrow$3.0)&99.9 ($\uparrow$4.5)\\
Llama3-8B-Instruct & 8B &12.6&17.3&47.3&&84.1\textsuperscript{$\dagger$}&82.4\textsuperscript{$\dagger$}&100
&&\underline{\textbf{84.7}}\textsuperscript{$\dagger$} ($\uparrow$0.6)& \underline{\textbf{83.7}}\textsuperscript{$\dagger$} ($\uparrow$1.3)&100\\ 
\hline
\end{tabular}
\label{tab:text_generation}
\end{subtable}
\begin{subtable}[h]{0.999\linewidth}
\begin{tabular}[]{lllllllllllll}
\multicolumn{13}{c}{\textbf{(b). \textit{SeqC}}} \\ \hline
&&\multicolumn{2}{c}{\textbf{NFT Baselines}} &
\multicolumn{9}{c}{\textbf{Fine-tuned Models}}\\ \cline{6-13}
&&&&&\multicolumn{2}{l}{\textcircled{\tiny 1} \textit{ natural input}}
&&\multicolumn{2}{l}{\textcircled{\tiny 2} \textit{ structured input}}
&&\multicolumn{2}{l}{\textcircled{\tiny 3} \textit{ inst + structured input}}\\
\cline{3-4} \cline{6-7} \cline{9-10} \cline{12-13}
base LM & size & acc. & m. f1 && acc. & m. f1 && acc. & m. f1 && acc. & m. f1  \\ \cline{3-4} \cline{6-7} \cline{9-10} \cline{12-13}
RoBERTa &125M&22.5&7.3 && 78.4 & 75.8 &&\underline{79.8} ($\uparrow$1.4)&\underline{78.4} ($\uparrow$2.6) &&78.8 ($\downarrow$1)&75.8 ($\downarrow$2.6)\\
GPT-j & 6B&16.0&11.2&&81.1&79.2 && 81.3 ($\uparrow$0.2)&80.0 ($\uparrow$0.8)&&\underline{82.2} ($\uparrow$0.9)&\underline{80.8} ($\uparrow$0.8) \\
Mistral-Instruct & 7B& 15.7 & 9.1 &&\underline{83.3}  &\underline{81.9} && 52.4 ($\downarrow$30.9) & 32.8 ($\downarrow$49.1) &&48.8 ($\downarrow$3.6)&32.4 ($\downarrow$0.4) \\
Llama2-7B & 7B & 22.4 & 14.1&&82.7&81.5& & 84.3 ($\uparrow$1.6) &\underline{83.3} ($\uparrow$1.8)  && \underline{84.5} ($\uparrow$0.2) & 83.0 ($\downarrow$0.3) \\
Llama2-7B-Chat & 7B & 24.2 &12.5&&81.6  &80.1  && \underline{84.4} ($\uparrow$2.8)& \underline{82.8} ($\uparrow$2.7)&& 83.8 ($\downarrow$0.6) & 82.1 ($\downarrow$0.7)  \\
Llama2-13B & 13B & 15.5 &5.4  &&84.0& 82.0& &84.9 ($\uparrow$0.9)  & 84.1 ($\uparrow$2.1)  && \underline{85.4}\textsuperscript{$\dagger$} ($\uparrow$0.5)  & \underline{\textbf{84.3}}\textsuperscript{$\dagger$} ($\uparrow$0.2) \\
Llama2-13B-Chat & 13B & 26.9 &13.0 && 83.0 &81.5 && 84.2 ($\uparrow$1.2) &82.5 ($\uparrow$1.0) && \underline{85.1} ($\uparrow$0.9) & \underline{83.7} ($\uparrow$1.2)  \\ 
Llama3-8B & 8B & 35.6 &13.0&&84.1 &82.3\textsuperscript{$\dagger$} & &\underline{84.2} ($\uparrow$0.1)& \underline{83.1} ($\uparrow$0.8)& & 46.8 ($\downarrow$37.4) & 26.4  ($\downarrow$56.7) \\
Llama3-8B-Instruct & 8B &10.6 &9.0& &84.4\textsuperscript{$\dagger$} &82.2&& \underline{\textbf{85.6}}\textsuperscript{$\dagger$} ($\uparrow$1.2)&\underline{\textbf{84.3}}\textsuperscript{$\dagger$} ($\uparrow$2.1)&&83.4 ($\downarrow$2.2)&81.9 ($\downarrow$2.4)\\ 
\bottomrule
\end{tabular}
\label{tab:seq_classification}
\end{subtable}
\caption[]{Results of human and instruction tuning baselines, approaches (a) \textit{Gen} and (b) \textit{SeqC}. Reported are accuracy (acc.), macro average F1 score (m. f1) and Answer Inclusion Rate (AIR) on the test set. For each base LM, we compare the performance of the non-fine-tuned model with that of models fine-tuned using different input formats, noting performance differences in parentheses.
The best-performing setting for each LM is underlined, and \textsuperscript{$\dagger$} denotes the best-performing LM within each setting. The best metrics from each approach are highlighted in bold.}
\label{tab:app_a_b}
\end{table*}

\subsection{Language Models}
\label{subsec:lms}
The proposed approaches are intended for systematically investigating fine-tuning LLMs but are readily extendable to other language models (LMs). We explore eight of the most advanced LLMs: GPT-j \cite{gpt-j}, Mistral-Instruct \cite{mistral}, Llama2-7B and Llama2-7B-Chat \cite{llama2}, Llama2-13B and Llama2-13B-Chat \cite{llama2}, Llama3-8B and Llama3-8B-Instruct\footnote{\url{https://github.com/meta-llama/llama3}}, and compare them with two PLMs: T5 \cite{t5} and RoBERTa \cite{roberta}. 
Details on model selection and an overview of the chosen LLMs and PLMs are provided in §\ref{sec:method_app}.

\begin{table*}[ht]
\fontsize{8}{8}
\selectfont
\centering
\renewcommand{\arraystretch}{1.2} 
\begin{tabular}[]{lllllllllllllll}
\hline
\multicolumn{15}{c}{\textbf{(c). \textit{SNet}}} \\\cline{2-15}
&\multicolumn{2}{l}{\textcircled{\tiny 1} \textit{ diff}}
&&\multicolumn{2}{l}{\textcircled{\tiny 2} \textit{ diffABS}}
&&\multicolumn{2}{l}{\textcircled{\tiny 3} \textit{ n-diffABS}}
&&\multicolumn{2}{l}{\textcircled{\tiny 4} \textit{ n-o}}
&&\multicolumn{2}{l}{\textcircled{\tiny 5} \textit{ n-diffABS-o}}
\\ \cline{2-3} \cline{5-6} \cline{8-9}\cline{11-12}\cline{14-15}
base LM  & acc. & m. f1 && acc. & m. f1 && acc. & m. f1&& acc. & m. f1&& acc. & m. f1\\ \cline{2-3} \cline{5-6} \cline{8-9}\cline{11-12}\cline{14-15}
Llama2-7B  &61.5&60.5&&\underline{69.7}&\underline{69.5}&&68.5&68.0&&60.8&58.8&&67.7&68.0\textsuperscript{$\dagger$} \\
Llama2-7B-Chat &60.7&56.5&&\underline{72.4}&\underline{71.4}&&65.4&64.7&&58.7&55.3&&68.5\textsuperscript{$\dagger$}&67.6 \\
Llama2-13B &62.4&59.3&&\underline{73.1}&\underline{72.4}&&67.5&67.2&&61.0\textsuperscript{$\dagger$}&59.1\textsuperscript{$\dagger$}&&66.0&67.2\\
Llama2-13B-Chat &63.7\textsuperscript{$\dagger$}&61.6\textsuperscript{$\dagger$}&&\underline{69.4}&\underline{69.3}&&66.9&66.3&&60.4&57.9&&66.0&65.3 \\ 
Llama3-8B &61.0&57.4&&\underline{70.6}&\underline{69.8}&&69.8\textsuperscript{$\dagger$}&68.7\textsuperscript{$\dagger$}&&58.6&56.6&&64.8&63.8 \\
Llama3-8B-Instruct &59.9&56.6&&\underline{\textbf{73.3}}\textsuperscript{$\dagger$}&\underline{\textbf{72.9}}\textsuperscript{$\dagger$}&&61.2&54.7&&60.6&58.4&&61.2&54.7 \\
\hline
\multicolumn{15}{c}{\textbf{(d). \textit{XNet}}} \\\cline{2-15}
&\multicolumn{2}{l}{\textcircled{\tiny 1} \textit{ diff}}
&&\multicolumn{2}{l}{\textcircled{\tiny 2} \textit{ diffABS}}
&&\multicolumn{2}{l}{\textcircled{\tiny 3} \textit{ n-diffABS}}
&&\multicolumn{2}{l}{\textcircled{\tiny 4} \textit{ n-o}}
&&\multicolumn{2}{l}{\textcircled{\tiny 5} \textit{ n-diffABS-o}}
\\ \cline{2-3} \cline{5-6} \cline{8-9}\cline{11-12}\cline{14-15}
base LM  & acc. & m. f1 && acc. & m. f1 && acc. & m. f1&& acc. & m. f1&& acc. & m. f1\\\cline{2-3} \cline{5-6} \cline{8-9}\cline{11-12}\cline{14-15}
Llama2-7B &83.0&81.4&&84.4&\underline{83.1}&&\underline{84.5}&82.8&&83.6&82.2&&83.2&81.6 \\
Llama2-7B-Chat &\underline{84.3}&\underline{83.2}&&83.6&81.9&&83.6&82.4&&83.3&81.4&&83.2&81.8 \\
Llama2-13B &84.3&82.7&&84.0&82.7&&\underline{85.0}&\underline{\textbf{83.9}}\textsuperscript{$\dagger$}&&84.4&83.4&&84.6\textsuperscript{$\dagger$}&83.7\textsuperscript{$\dagger$}\\
Llama2-13B-Chat &
84.3&82.9&&\underline{\textbf{85.2}}\textsuperscript{$\dagger$}&\underline{83.7}\textsuperscript{$\dagger$}&&84.5&83.6&&84.9&83.7\textsuperscript{$\dagger$}&&84.6\textsuperscript{$\dagger$}&83.3\\ 
Llama3-8B &83.7&82.4&&84.1&82.4&&\underline{84.7}&\underline{83.6}&&76.7&73.7&&83.5&82.1 \\
Llama3-8B-Instruct &84.4\textsuperscript{$\dagger$}&83.4\textsuperscript{$\dagger$}&&84.5&83.2&&\underline{85.1}\textsuperscript{$\dagger$}&\underline{83.7}&&\underline{85.1}\textsuperscript{$\dagger$}&\underline{83.7}\textsuperscript{$\dagger$}&&84.1&83.3 \\
\bottomrule
\end{tabular}
\caption[LLMs]{Results of approaches (c) \textit{SNet} and (d) \textit{XNet}. Reported are accuracy (acc.) and macro average F1 score (m. f1) on the test set. For each base LM, we compare the performance of models fine-tuned using different transformation functions (§\ref{subsec:trans_func}).
The best-performing setting for each LM is underlined, \textsuperscript{$\dagger$} denotes the best-performing LM within each setting. 
The best metrics from each approach are in bold.
}
\label{tab:app_c_d}
\end{table*}

\section{Results and Discussion}
\label{sec:results}
\subsection{Data and Experimental Details}
\label{subsec:data}
For our experiments, we seek a high-quality dataset with a sufficient number of samples for fine-tuning.
Re3-Sci \cite{re3} is such a dataset, which comprises 11,566 high-quality human-labeled sentence edits from 314 document revisions. We divide the dataset into training, validation, and test sets with 7,478/1,776/2,312 edits.
Re3-Sci categorizes edit intents into five distinct labels: \textit{Grammar} and \textit{Clarity} for surface language improvements, \textit{Fact/Evidence} and \textit{Claim} for semantic changes in factual content or statements, and \textit{Other} for all other cases. The task is thus formulated as a 5-class classification challenge given a sentence revision pair (§\ref{subsec:approaches}). 
We fine-tune all linear layers of the LLMs using QLoRA \cite{qlora}. The PLMs are fully fine-tuned with all weights being directly updated. 
For approach \textit{Gen}, the output token limit is set to ten. We define \textit{Answer Inclusion Rate (AIR)} as the percentage of samples where a label string falls within the ten output tokens, regardless of correctness. 
Further details are provided in §\ref{sec:exp_details}.

\subsection{Discussion}
\label{subsec:diss}
Table \ref{tab:app_a_b} shows the performance of human annotators and instruction tuning baselines using GPT-4 and Llama2-70B (details in §\ref{sec:exp_details}), as well as the performance from approaches \textit{Gen} and \textit{SeqC}, comparing various input types. 
Table \ref{tab:app_c_d} presents the comparative results of approaches \textit{SNet} and \textit{XNet}, evaluating different transformation functions. Based on these results, we address five research questions:

\noindent \textbf{RQ1: Are fine-tuned LLMs good edit intent classifiers compared to fully fine-tuned PLMs and instruction-tuned larger LLMs?} 
Our results suggest that LLMs can be effectively enhanced to serve as good edit intent classifiers with our optimal approaches, outperforming larger instruction-tuned LLMs and fully fine-tuned PLMs, and achieving new state-of-the-art (SOTA) performance on the Re3-Sci dataset.
First, we compare our best results with the baselines. 
Bold texts in Table \ref{tab:app_a_b}(b) indicate that approach \textit{SeqC} with either Llama2-13B or Llama3-8B-Instruct achieves the highest macro average F1 score of 84.3.
This result notably exceeds the GPT-4 baselines, both in a zero-shot setting and when enhanced with ICL and CoT. It also substantially surpasses the previous SOTA results achieved by an instruction-tuned Llama2-70B, as reported by \citet{re3}.
Then, we compare the results from fine-tuning LLMs and PLMs. 
Table \ref{tab:app_a_b}(b) shows that using the encoding-based approach \textit{SeqC}, most of the eight LLMs outperform a fully fine-tuned RoBERTa across various input formats, highlighting the superior encoding capabilities of LLMs.
Table \ref{tab:app_a_b}(a) shows that using approach \textit{Gen} with structured inputs, Llama2-13B-Chat, Llama3-8B, and Llama3-8B-Instruct can achieve better or comparable results to a fully fine-tuned T5.
The favorable results in Table \ref{tab:app_c_d}(d) indicate that fine-tuning via \textit{XNet} also effectively enhances LLMs as edit intent classifiers. 

\noindent \textbf{RQ2: Which LLMs are more effective as edit intent classifiers?} Overall, an analysis of the best-performing fine-tuned models, marked with \textsuperscript{$\dagger$} in Tables \ref{tab:app_a_b} and \ref{tab:app_c_d}, reveals that the 13B Llama2 and 8B Llama3 models demonstrate the greatest potential and achieve the best results.
Additionally, we observe that using the \textit{Gen} approach (Table \ref{tab:app_a_b}(a)), instruction-fine-tuned LLMs consistently outperform their non-instruction-fine-tuned counterparts, with statistical significance supported by paired one-sided two-sample t-tests \cite{ttest} and one-sided Wilcoxon signed-rank tests  \cite{wilcoxon}. 
This performance improvement is likely attributable to the enhanced capability of instruction-fine-tuned models to comprehend complex task instructions and label tags.
However, in \textit{SeqC}, \textit{SNet} and \textit{XNet} approaches, there are no consistent performance differences between the chat and non-chat versions of LLMs.

\begin{figure}[t]
  \centering
  \includegraphics[width=0.33\textwidth]{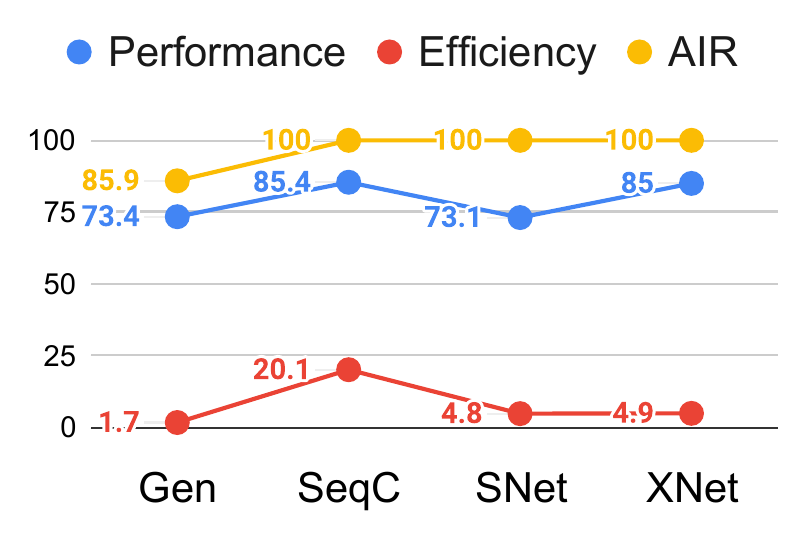}
  \caption{Approaches comparison using Llama2-13B as the base language model. AIR: Answer Inclusion Rate; performance: accuracy; efficiency: the number of samples processed per second during inference.}
  \label{fig:approaches_comp}
\end{figure}

\noindent \textbf{RQ3: Which approach is most effective?} 
Overall, the \textit{SeqC} approach demonstrates superior performance, answer inclusion rate (AIR), and inference efficiency.
Table \ref{tab:app_a_b}(a) indicates that generative models encounter AIR issues even after fine-tuning, suggesting that the generation-based approach is not optimal in practice due to its lack of robustness and difficulty in control.
In terms of performance, approaches \textit{SeqC} and \textit{XNet} are superior. The cross network (\textit{XNet}) consistently and substantially outperforms the Siamese network (\textit{SNet}) when using the same LLMs and transformation functions (Table \ref{tab:app_c_d}). 
The \textit{SeqC} approach demonstrates notable superiority in inference efficiency, measured by the number of samples processed per second during inference, making it particularly well-suited for application to large datasets.
Figure \ref{fig:approaches_comp} compares the four approaches across the three aspects, using Llama2-13B as the base language model. The \textit{SeqC} approach achieves perfect AIR, the best performance, and a 12x inference speedup compared to the \textit{Gen} approach and a 4x speedup compared to \textit{SNet} and \textit{XNet}. 

\noindent \textbf{RQ4: What are the effects of the input types?}
Now, we examine the ablation results detailed in parentheses in Table \ref{tab:app_a_b}. 
Table \ref{tab:app_a_b}(a) shows that using structured input instead of natural language input improves performance for the Llama2 models in approach \textit{Gen}, though it may decrease AIR. However, for GPT-j and Mistral-Instruct, structured input has a substantial negative impact.
Table \ref{tab:app_a_b}(b) shows that in approach \textit{SeqC}, using structured inputs positively impacts RoBERTa and all LLMs except for Mistral-Instruct. Adding the task instruction to structured inputs has minimal effects on most models, however, it particularly negatively impacts Llama3-8B.

\noindent \textbf{RQ5: What are the effects of the transformation functions?} 
We examine the most effective transformation functions, indicated by the most frequently underlined columns in Table \ref{tab:app_c_d}. 
Table \ref{tab:app_c_d}(c) indicates that when using \textit{SNet}, $f_{diffABS}$ substantially outperforms all other functions across all LLMs.
For \textit{XNet}, the best-performing functions are $f_{n-diffABS}$ and $f_{diffABS}$, as indicated in Table \ref{tab:app_c_d}(d).
These results align with our intuition that the EIC task focuses on analyzing variations between text versions, and incorporating sentence embedding differences proves to be effective.

\subsection{Generalization Evaluation}
\label{subsec:gen_eval}
We assess the generalization of our findings from the EIC task across five additional classification tasks from the MTEB benchmark \cite{mteb}. The selected tasks comprise three binary pair classification tasks, in which the inputs are sentence pairs and the outputs are binary labels, along with two multi-class single-input classification tasks, where the inputs consist of individual sentences and the output labels are multi-class. The former group features a similar input architecture to that of EIC, allowing for the application of all four approaches. The latter type exhibits output complexity similar to that of EIC, featuring multiple potential labels.

\textbf{Our evaluations across the five tasks provide compelling evidence that:} (1) LLMs can be fine-tuned to operate as good classifiers, achieving SOTA results on the additional tasks; (2) among the eight tested LLMs, the 13B Llama2 and 8B Llama3 models exhibit the greatest potential and achieve best results; and (3) the encoding-based \textit{SeqC} approach proves to be the most effective, demonstrating significantly superior performance, inference efficiency, and perfect AIR.
Appendix §\ref{sec:gen} gives details on the datasets and tasks, experimental settings and results, as well as further discussion. 

\section{Application: \textit{Re3-Sci2.0}}
\label{sec:auto_anno}

The original Re3-Sci dataset contains only 314 documents covering limited research domains, thus constraining in-depth science-of-science analysis of how humans improve scientific quality through revisions and how their document-based editing behavior varies across domains. Having determined the optimal approach for EIC among the considered ones, we apply our best-performing model to create \textit{Re3-Sci2.0}: the first large-scale corpus of academic document revisions for edit analysis across research domains.

\subsection{Data Collection and Labeling}
Re3-Sci is built upon F1000RD \citep{f1000rd} and the ARR-22 subset of NLPeer \citep{nlpeer}, which include revisions of scientific papers and associated reviews.
We extend the Re3-Sci dataset by annotating the remaining documents from the two source corpora totaling 1,780 scientific document revisions: 325 from NLPeer and 1,455 from F1000RD.

The automatic annotation consists of two steps: (1). \textbf{Revision Alignment (RA)} to identify sentence revision pairs as well as additions and deletions of sentences, and label them with action labels "Modify", "Add" or "Delete". 
We fine-tune a Llama2-13B classifier using \textit{SeqC} achieving an accuracy of 99.3\%, and employ a two-stage method as detailed in §\ref{subsec:app_rev_align}.
(2). \textbf{EIC} to label the identified edits with intent labels.
We use the best-performing Llama2-13B\footnote{We did not use the Llama3 classifiers since Llama3 was released after our auto-annotation process was completed.} classifier (§\ref{sec:results}), as it achieves the best performance, perfect AIR and high inference efficiency. 
A human evaluation of 10 randomly selected documents with 348 edits reveals 100\% accuracy for RA and 90.5\% accuracy for EIC (details in §\ref{subsec:app_human_eval}).

\subsection{Basic Statistics and Subsets}
The \textit{Re3-Sci2.0} dataset includes 1,780 document revisions with 94,482 edits, each annotated with action and intent labels. 
The 325 documents from NLPeer are all from the NLP field (\textit{nlp}), whereas the documents from F1000RD fall into three main subject domains: Natural Sciences (\textit{nat}), Medical and Health Sciences (\textit{med}) and Social Sciences (\textit{soc}). 
Specific documents from the medical domain that provide brief reports on individual medical cases are separated from standard medical research papers to form a distinct \textit{case} category.
Similarly, documents from the natural sciences domain that provide technical reports on software or tools, primarily from computational biology, are separated into the \textit{tool} category.
§\ref{subsec:app_area_cat} provides detailed definitions of the research domains and document categories, Table \ref{tab:re3_v2_stat} presents statistics for each subset.

\begin{table}[ht]
\fontsize{8}{8}
\selectfont
\centering
\renewcommand{\arraystretch}{1.2} 
\begin{tabular}[]{llllll}
\toprule
&doc. & edit & d\_word & d\_sent. &d\_edit \\\hline
all &1,780 &94,482&4,650&201&53\\\hline
nlp&325&29,782&5,775&262&92\\
case (med)&112&2,248&2,118&100&20\\
med&208&7,521&4,616&193&36\\
tool (nat)&162&7,143&3,505&170&44\\
nat&349&18,834&5,001&210&54\\
soc&46&2,466&4,888&206&54\\
\bottomrule
\end{tabular}
\caption[]{\textit{Re3-Sci2.0} statistics and subsets. Presented are counts of documents and total sentence edits, and average counts of words, sentences and edits per document.}
\label{tab:re3_v2_stat}
\end{table}

\subsection{Analysis of Editing Behavior}
\label{sec:meta_analysis}
As a resource, \textit{Re3-Sci2.0} enables new empirical insights into the text editing behavior in the academic domain. We illustrate this analysis by investigating the following research questions:

\noindent \textbf{RQ1: How do successful revisions enhance scientific quality compared to unsuccessful ones?}
We interpret increased review scores between document versions as indicators of successful revisions and improvements in scientific quality (more details in §\ref{subsec:app_success_fail}).
We investigate the focus of authors' revisions by analyzing the document-based proportions of edit action and intent combinations as key variables.
A value of 1 is assigned to successfully revised documents with increased review scores and 0 to unsuccessful ones. We then fit a binary logistic regression model to predict revision success, which is statistically significant with an LLR p-value of 0.001. Table \ref{tab:success_reg} shows that focusing on modifications to enhance clarity and claims, and additions of new facts or evidence, significantly and positively influences the success of revisions.
Additionally, Table \ref{tab:frd_success_fail_count} in §\ref{subsec:app_success_fail} indicates that successful revisions include significantly more edits compared to unsuccessful ones.
\begin{table}[t]
\fontsize{8}{8}
\selectfont
\centering
\renewcommand{\arraystretch}{1.2} 
\begin{tabular}[]{lllllll}
\toprule 
&\textit{coef}&\textit{p-value}\\\hline
Add, Fact/Evidence&\textbf{0.9341}&0.003\\
Add, Claim&0.6116&0.221\\
Delete, Fact/Evidence&2.0920&0.061\\
Delete, Claim&2.9626&0.076\\
Modify, Grammar&-0.5324&0.161\\
Modify, Clarity&\textbf{1.0723}&0.004\\
Modify, Fact/Evidence&0.3506&0.347\\
Modify, Claim&\textbf{3.3392}&0.040\\
\bottomrule
\end{tabular}
\caption[]{Results of the binary logistic regression.
Presented are the regression coefficients for the variables. Bold values indicate statistical significance (p < 0.05).}
\label{tab:success_reg}
\end{table}

\begin{figure*}[ht]
  \centering
  \begin{subfigure}[b]{0.325\columnwidth}
    \includegraphics[width=\linewidth]{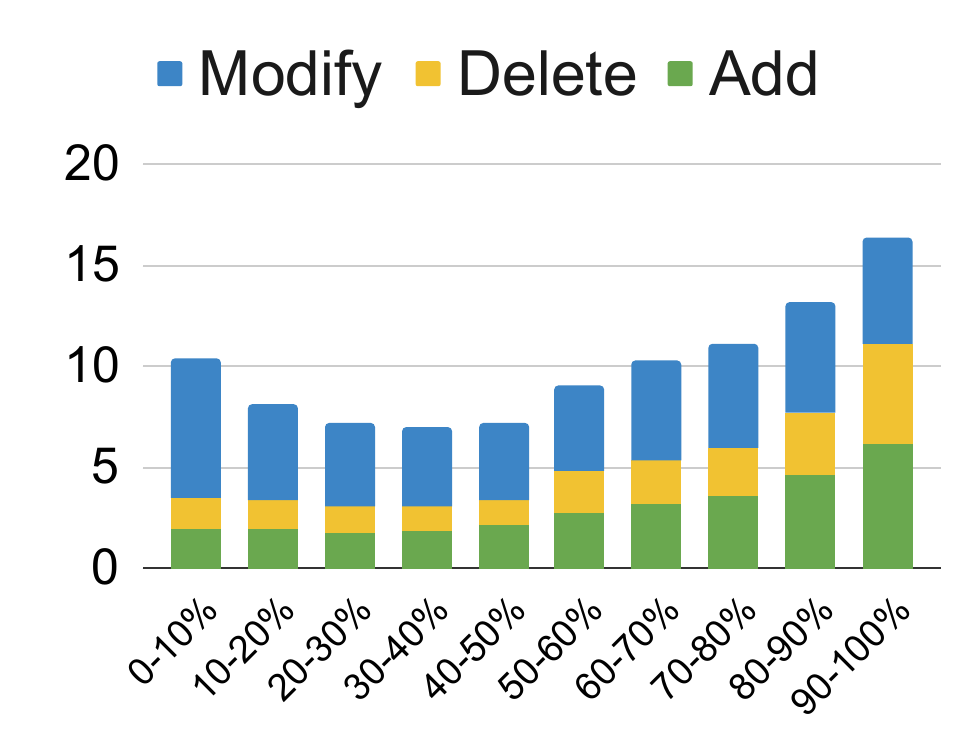}
    \caption{nlp: action}
    \label{fig:loc_ea_nlp}
  \end{subfigure}
  \begin{subfigure}[b]{0.325\columnwidth}
    \includegraphics[width=\linewidth]{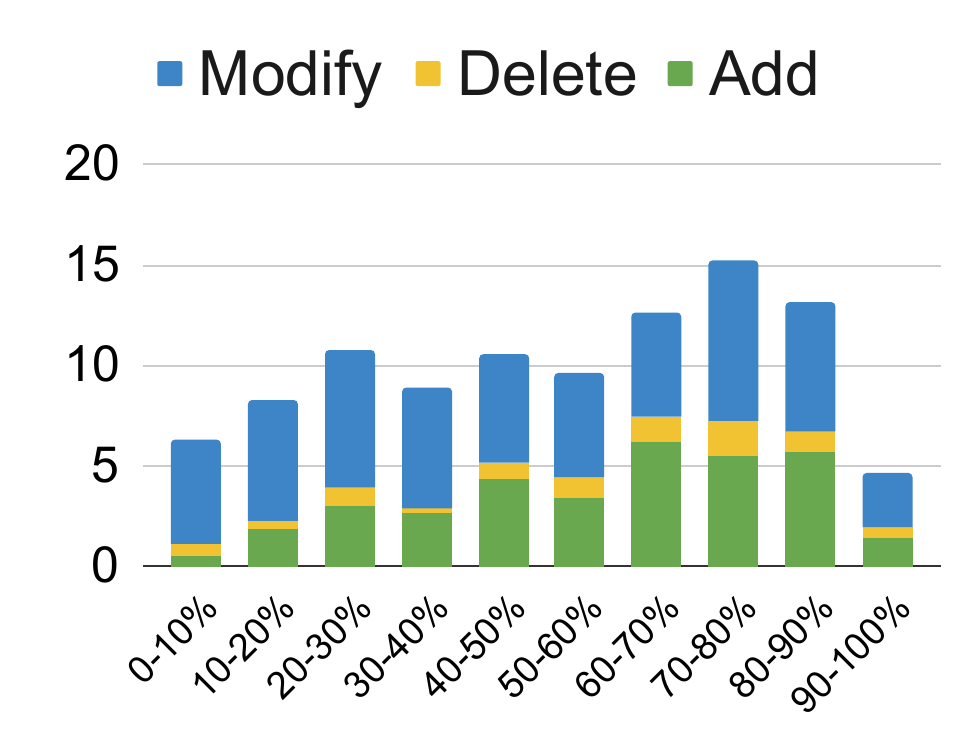}
    \caption{case: action}
    \label{fig:loc_ea_case}
  \end{subfigure}
  \begin{subfigure}[b]{0.325\columnwidth}
    \includegraphics[width=\linewidth]{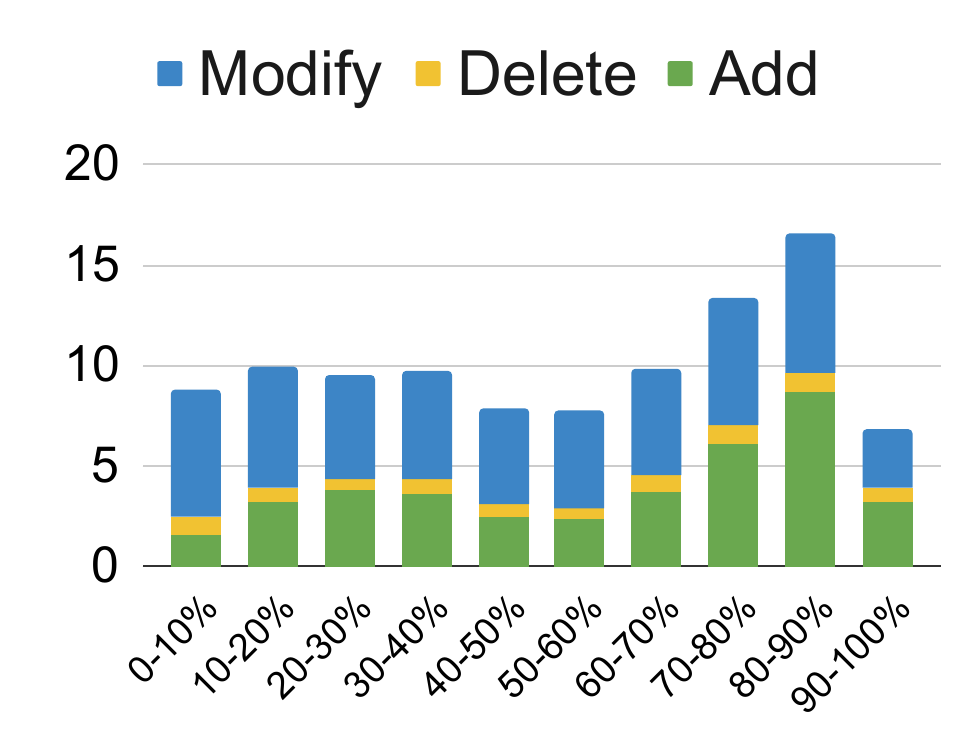}
    \caption{med: action}
    \label{fig:loc_ea_med}
  \end{subfigure}
  \begin{subfigure}[b]{0.325\columnwidth}
    \includegraphics[width=\linewidth]{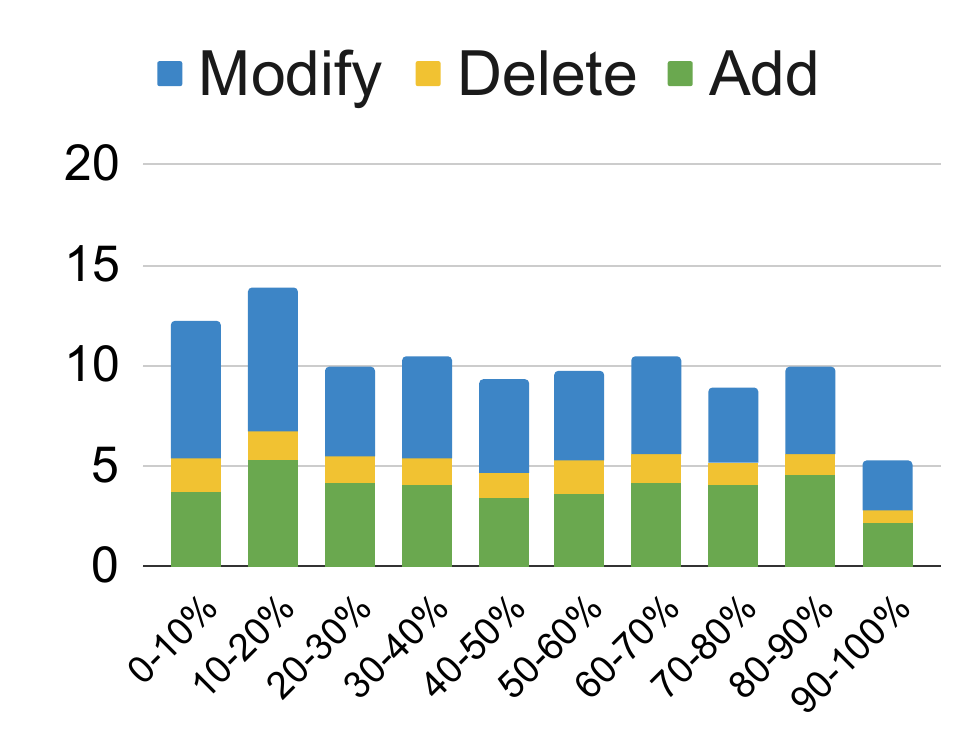}
    \caption{tool: action}
    \label{fig:loc_ea_tool}
  \end{subfigure}
  \begin{subfigure}[b]{0.325\columnwidth}
    \includegraphics[width=\linewidth]{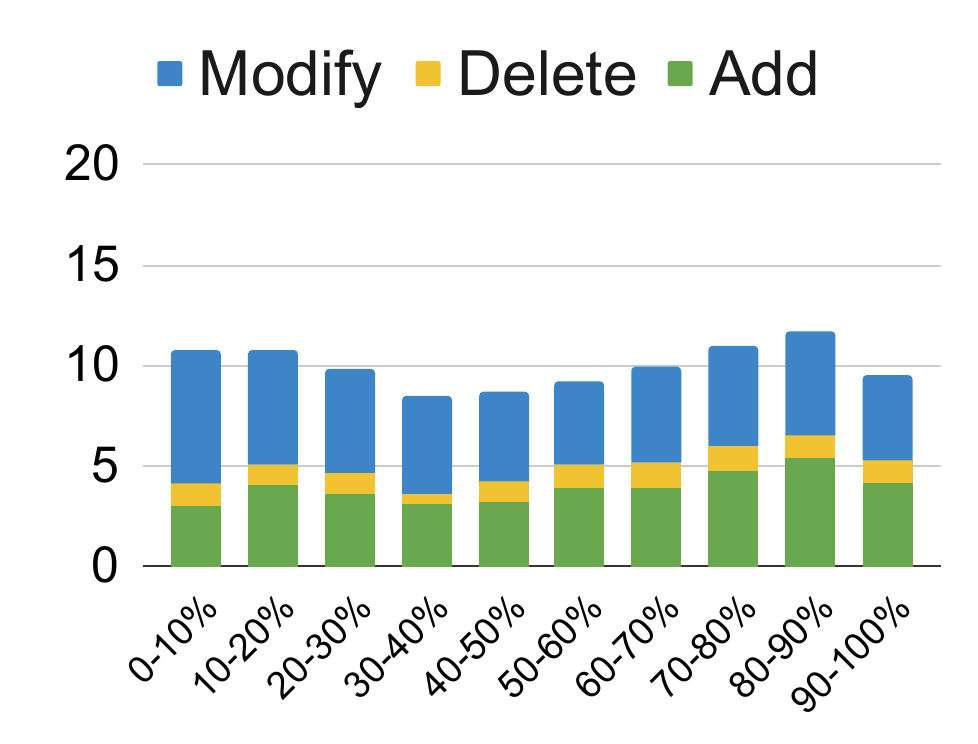}
    \caption{nat: action}
    \label{fig:loc_ea_nat}
  \end{subfigure}
  \begin{subfigure}[b]{0.325\columnwidth}
    \includegraphics[width=\linewidth]{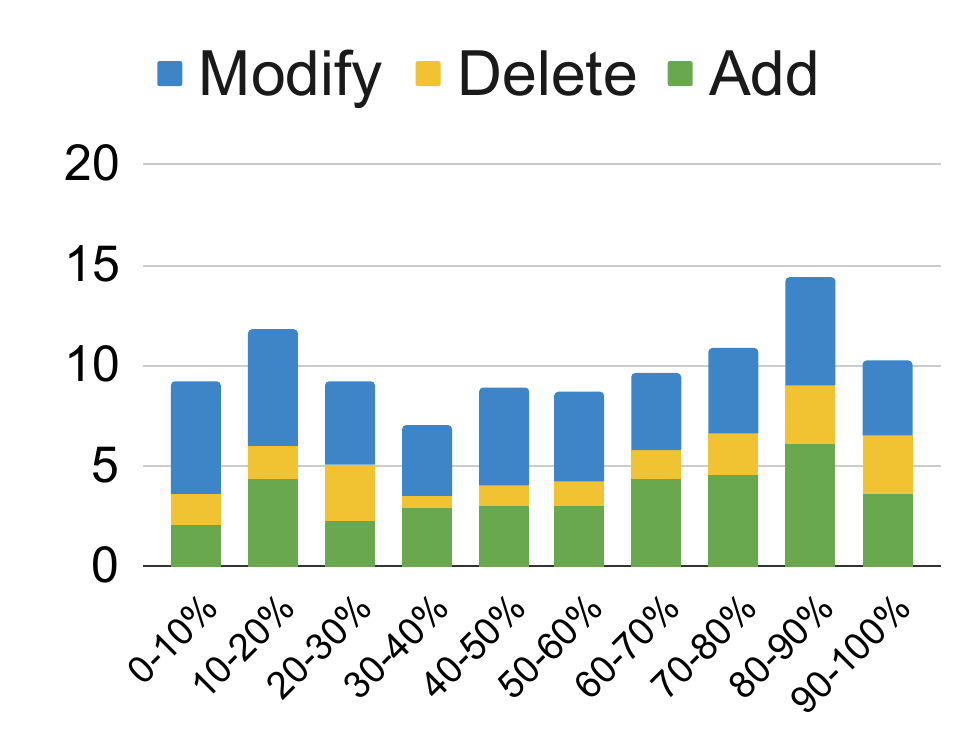}
    \caption{soc: action}
    \label{fig:loc_ea_soc}
  \end{subfigure}\\
  \begin{subfigure}[b]{0.325\columnwidth}
    \includegraphics[width=\linewidth]{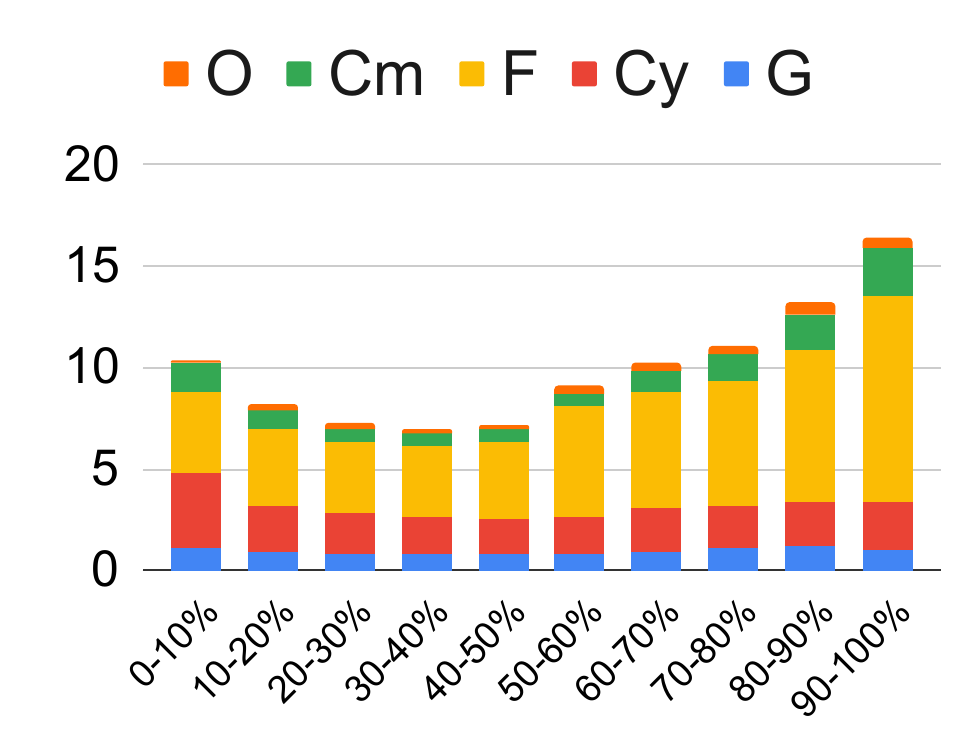}
    \caption{nlp: intent}
    \label{fig:loc_ei_nlp}
  \end{subfigure}
  \begin{subfigure}[b]{0.325\columnwidth}
    \includegraphics[width=\linewidth]{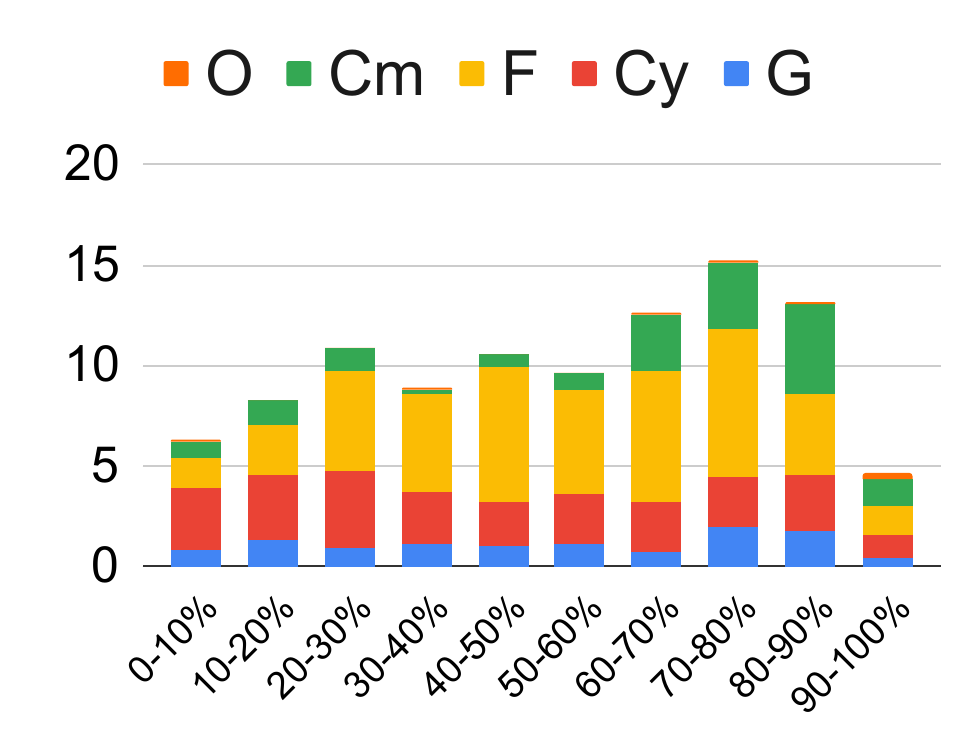}
    \caption{case: intent}
    \label{fig:loc_ei_case}
  \end{subfigure}
  \begin{subfigure}[b]{0.325\columnwidth}
    \includegraphics[width=\linewidth]{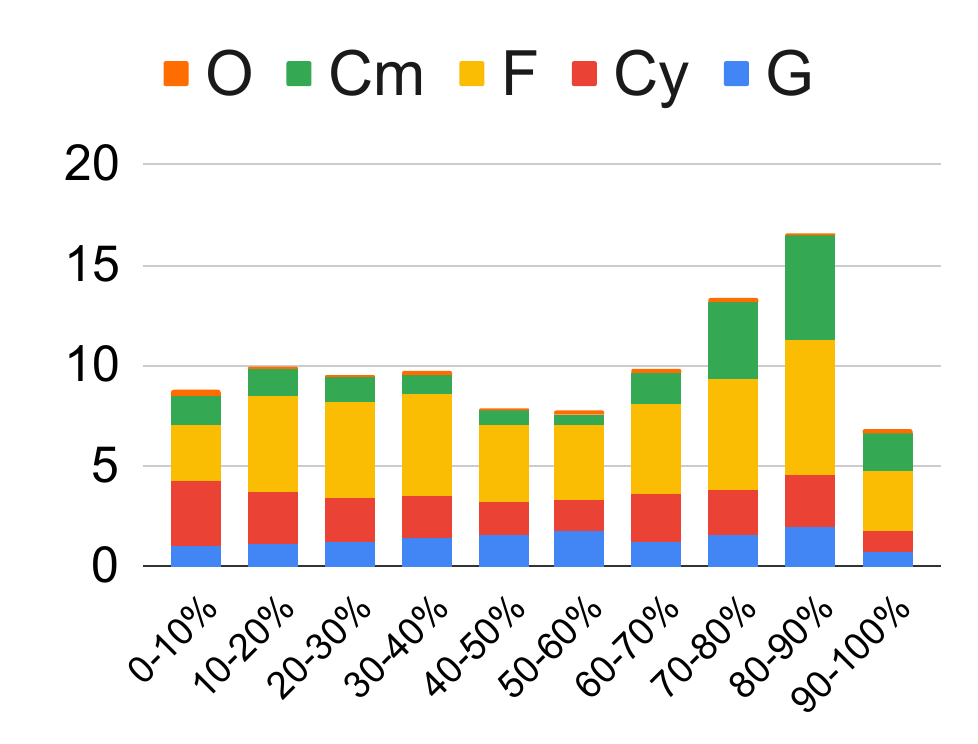}
    \caption{med: intent}
    \label{fig:loc_ei_med}
  \end{subfigure}
  \begin{subfigure}[b]{0.325\columnwidth}
    \includegraphics[width=\linewidth]{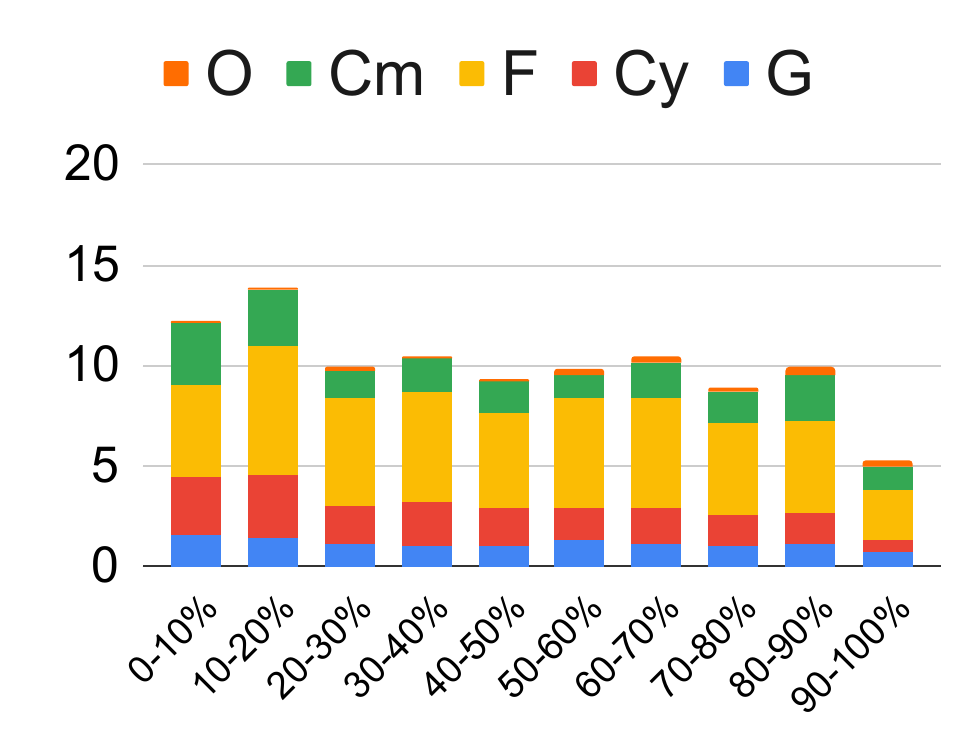}
    \caption{tool: intent}
    \label{fig:loc_ei_tool}
  \end{subfigure}
  \begin{subfigure}[b]{0.325\columnwidth}
    \includegraphics[width=\linewidth]{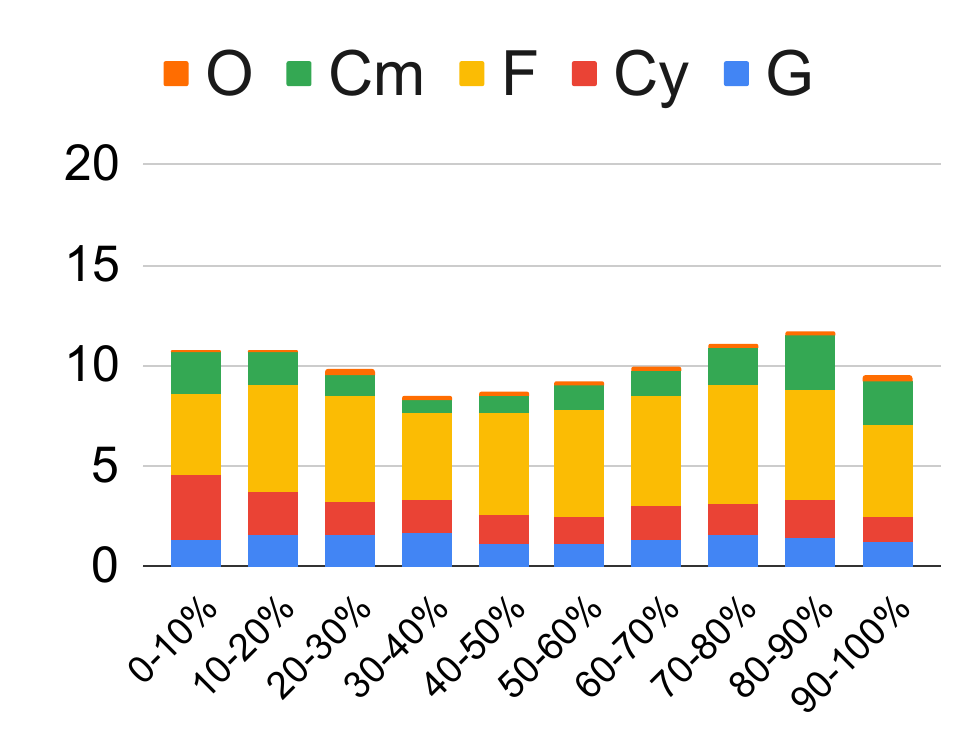}
    \caption{nat: intent}
    \label{fig:loc_ei_nat}
  \end{subfigure}
  \begin{subfigure}[b]{0.325\columnwidth}
    \includegraphics[width=\linewidth]{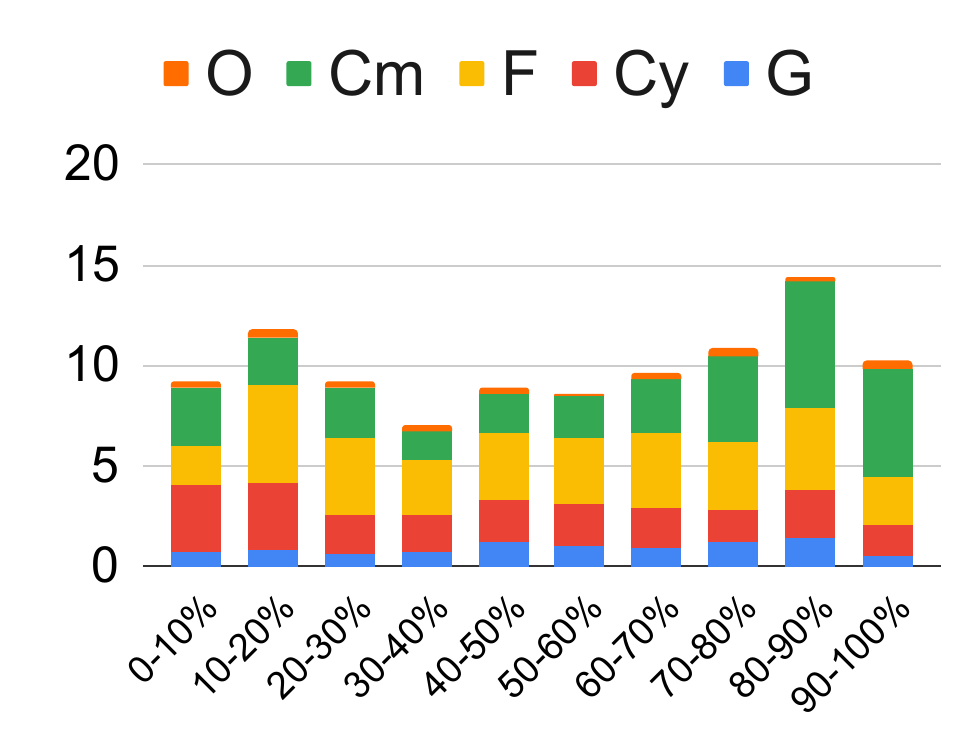}
    \caption{soc: intent}
    \label{fig:loc_ei_soc}
  \end{subfigure}\\
  \caption{Edit action and intent labels distribution over documents. The x-axis represents the relative sentence
positions within documents. G: Grammar, Cy: Clarity, F: Fact/Evidence, Cm: Claim, O: Other.}
  \label{fig:loc}
\end{figure*}

\begin{figure}[ht]
  \centering
  \begin{subfigure}[b]{0.325\columnwidth}
    \includegraphics[width=\linewidth]{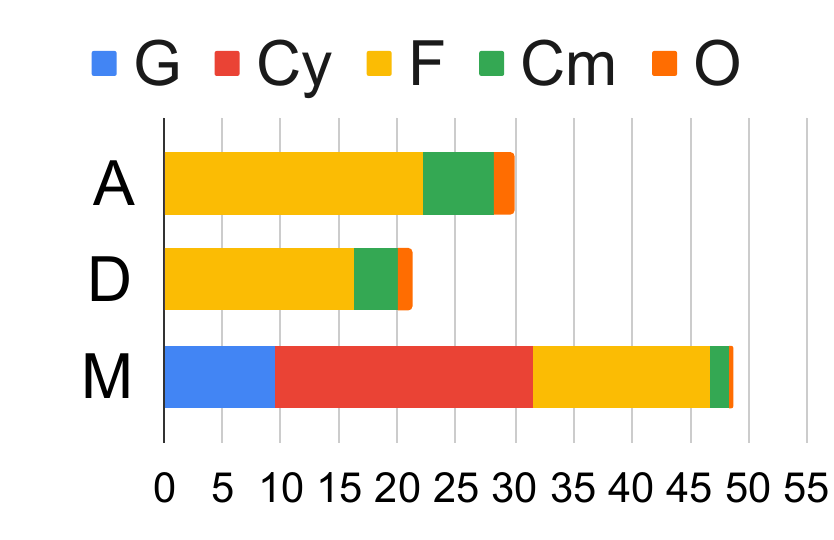}
    \caption{nlp}
    \label{fig:ea_ei_nlp}
  \end{subfigure}
  \begin{subfigure}[b]{0.325\columnwidth}
    \includegraphics[width=\linewidth]{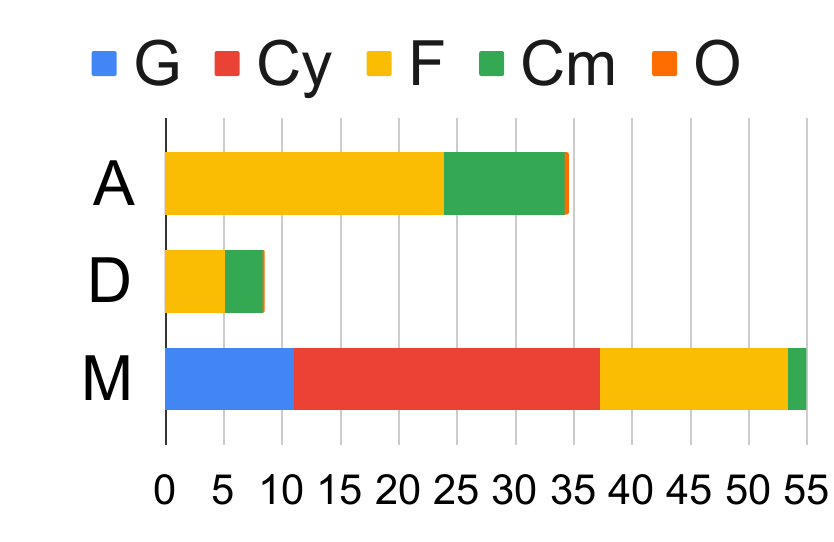}
    \caption{case}
    \label{fig:ea_ei_case}
  \end{subfigure}
  \begin{subfigure}[b]{0.325\columnwidth}
    \includegraphics[width=\linewidth]{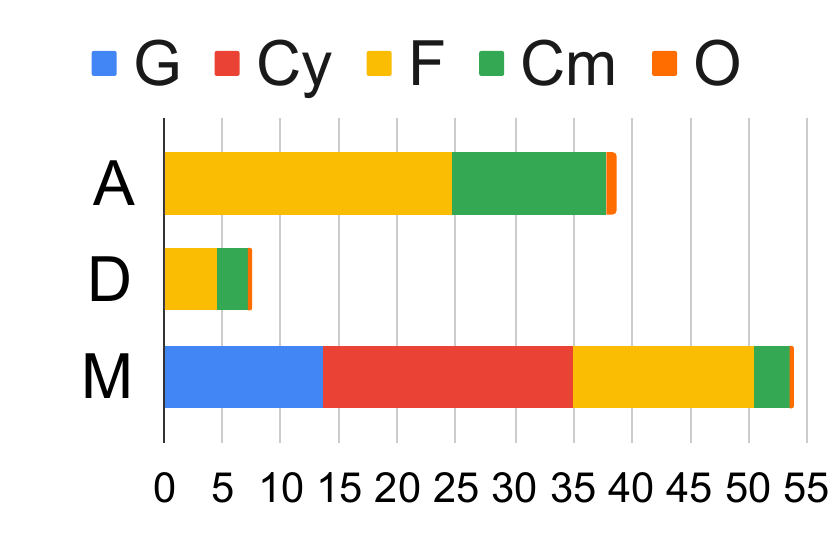}
    \caption{med}
    \label{fig:ea_ei_med}
  \end{subfigure} \\
  \begin{subfigure}[b]{0.325\columnwidth}
    \includegraphics[width=\linewidth]{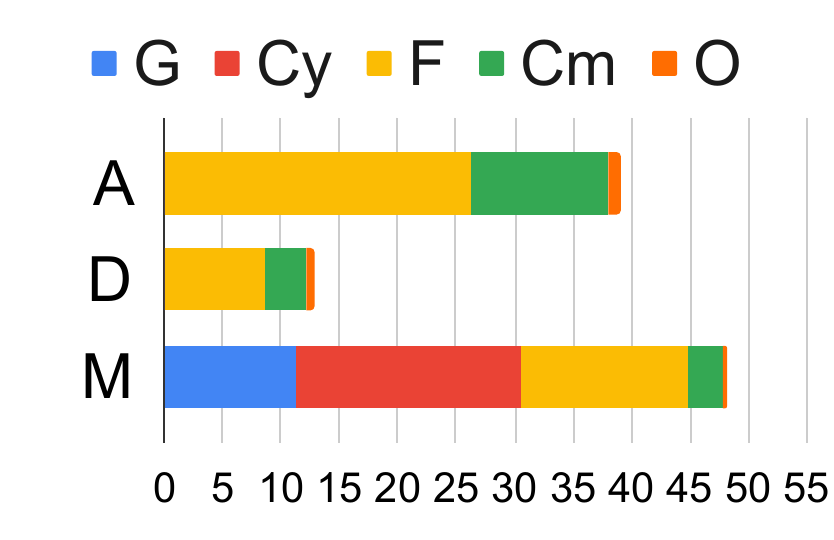}
    \caption{tool}
    \label{fig:ea_ei_tool}
  \end{subfigure}
  \begin{subfigure}[b]{0.325\columnwidth}
    \includegraphics[width=\linewidth]{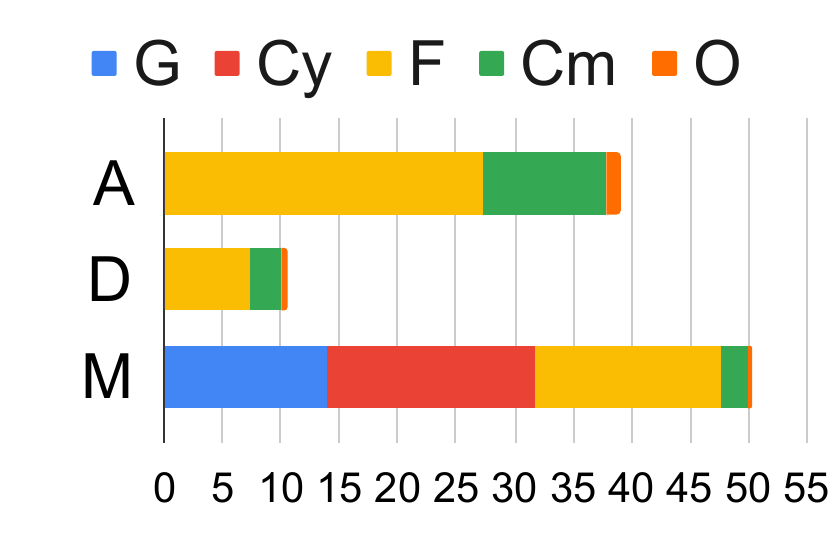}
    \caption{nat}
    \label{fig:ea_ei_nat}
  \end{subfigure}
  \begin{subfigure}[b]{0.325\columnwidth}
    \includegraphics[width=\linewidth]{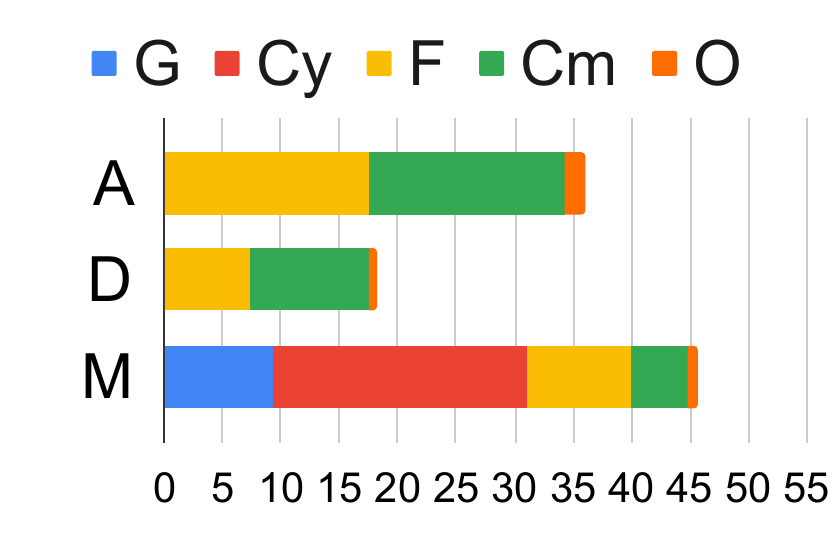}
    \caption{soc}
    \label{fig:ea_ei_soc}
  \end{subfigure} \\
  \caption{Combinations of edit action and intent labels across various categories. A: Add, D: Delete, M: Modify, G: Grammar, Cy: Clarity, F: Fact/Evidence, Cm: Claim, O: Other.}
  \label{fig:ea_and_ei}
\end{figure}

\noindent \textbf{RQ2: How do human editing behaviors differ across various research domains and document categories?}
To analyze human editing behaviors, we examine the proportions of action and intent combinations to reflect authors' editing focus (Figure \ref{fig:ea_and_ei}) and analyze the distribution of edits across documents to identify editing location (Figure \ref{fig:loc}). A Kullback–Leibler Divergence (KL) analysis of the distributions across research domains and document categories is shown in Figure \ref{fig:KL} in §\ref{subsec:app_rev_area}.

\textbf{Analysis indicates that human editing behaviors are consistent within the same research domain, despite variations in document categories.} For example, consider the \textit{case} and \textit{med} categories, both from the medical domain.
Table \ref{tab:re3_v2_stat} shows that medical case reports (\textit{case}) are generally shorter with fewer edits compared to other documents in the medical sciences (\textit{med}).
However, the revision focus of the authors appears similar, as illustrated in Figure \ref{fig:ea_ei_case} and Figure \ref{fig:ea_ei_med}. This similarity is further substantiated by the low KL values between \textit{case} and \textit{med} shown in Figure \ref{fig:ea_ei_kl} in §\ref{subsec:app_rev_area}. 
The revision locations for both action and intent in \textit{case} and \textit{med} are also similar, as evidenced by comparing Figure \ref{fig:loc_ea_case} and Figure \ref{fig:loc_ea_med}, as well as Figure \ref{fig:loc_ei_case} and Figure \ref{fig:loc_ei_med}. These similarities are supported by low KL scores between \textit{case} and \textit{med} in both Figure \ref{fig:ea_loc_kl} and Figure \ref{fig:ei_loc_kl}. 
Similarly, when comparing \textit{tool} and \textit{nat} across Figures \ref{fig:loc}, \ref{fig:ea_and_ei} and \ref{fig:KL}, it is evident that human editing focus and location are consistent within the natural sciences, regardless of different document categories.

Regarding \textbf{editing focus}, Figure \ref{fig:ea_and_ei} 
indicates that authors in the medical domain (\textit{case} and \textit{med}) and natural sciences (\textit{tool} and \textit{nat}) tend to make fewer deletions. In contrast, authors in NLP (\textit{nlp}) and social sciences (\textit{soc}) make more deletions, with the former emphasizing Fact/Evidence and the latter focusing more on Claim.  
Figure \ref{fig:ea_ei_kl} further shows that the social sciences domain differs most substantially from other domains in terms of editing focus, as indicated by the high KL scores between \textit{soc} and other domains.
Regarding \textbf{editing location}, Figure \ref{fig:loc} illustrates that in NLP, the final parts of documents are most frequently revised, primarily through additions and deletions of Fact/Evidence and Claim. In medical sciences (\textit{case} and \textit{med}), the 70-90\% range of relative document positions is intensively revised, characterized by more additions and claim changes compared to other locations. In natural sciences (\textit{tool} and \textit{nat}) and social sciences (\textit{soc}), edits tend to be more evenly distributed.

\section{Conclusion}
We have introduced a general framework for fine-tuning LLM classifiers, including four approaches, various LLM families, and training strategies. 
Extensive experiments on EIC have demonstrated that LLMs can be effectively fine-tuned as intent classifiers, outperforming fully fine-tuned PLMs and achieving SOTA results. Among the four approaches, the encoding-based \textit{SeqC} approach has shown superiority in model performance, inference efficiency, and answer inclusion. Furthermore, we have demonstrated the versatility of our framework and evaluated the generalizability of our findings on five additional classification tasks.

Using the best-performing  EIC model, we have annotated a large-scale dataset of scientific document revisions, enabling in-depth empirical analysis of revision success and human editing behavior across various research domains. 
Our illustratory analysis suggests that (1) focus on Clarity and Claim modifications and Fact/Evidence additions significantly and positively impacts revisions success; (2) human editing focus and location remain consistent within the same research domain regardless of document categories but vary substantially across different domains.

Our work paves the way for systematic investigation of LLMs for classification tasks and beyond. The general experimental framework is applicable to a wide range of classification tasks. The new dataset provides a robust foundation for multifaceted research in human editing in scientific domain and beyond. The annotation models and the labeling process are reusable and can be applied to generate new high-quality, large-scale automatically labeled revision datasets, as more raw data becomes available.

\section*{Limitations}
This study has several limitations that should be considered when interpreting the results. 
From a task and modeling perspective, this work focuses on edit intent classification, aiming to address this complex, challenging, yet underexplored task and facilitate crucial but understudied real-world applications for science-of-science analysis.  
While we conducted extensive generalization evaluations, the experimental results and discussions may not be directly applicable to other classification tasks. However, the proposed approaches and training strategies can be readily adapted to other classification tasks within our experimental framework.

From a data and analysis standpoint, the study's focus on English-language scientific publications stems from the limited availability of openly licensed scholarly publications in other languages. The use of Re3-Sci is driven by the need for high-quality and sufficiently large datasets for fine-tuning. 
Exploring the transferability of our findings to new languages, domains, and editorial workflows represents a promising direction for future research. When new data becomes available, our publicly available models can be used for annotation and analysis. Additionally, our experimental framework facilitates easy fine-tuning on other datasets and allows for systematic comparisons of various approaches and training strategies.

Finally, we highlight that our analysis serves an illustrative purpose. Its primary goal is to inspire researchers from other related disciplines to use natural language processing to explore new questions about editing, academic publishing and communication. Enabled by the new dataset and methods, we leave the in-depth investigation of human editing behavior across research communities for future research.

\section*{Ethics Statement} 
Re3-Sci and both subsets of the source data are licensed under CC-BY-NC 4.0, ensuring that the construction and use of our dataset comply with licensing terms. Our annotated dataset is available under a CC-BY-NC 4.0 license. The automatic annotation and analysis process does not involve the collection of any personal or sensitive information.
For privacy protection, author metadata has been omitted from the data release.

\section*{Acknowledgements}
This work is part of the InterText initiative\footnote{\url{https://intertext.ukp-lab.de/}} at the UKP Lab. This work has been funded by the German Research Foundation (DFG) as part of the PEER project (grant GU 798/28-1) and co-funded by the LOEWE Distinguished Chair “Ubiquitous Knowledge Processing”, LOEWE initiative, Hesse, Germany (Grant Number: LOEWE/4a//519/05/00.002(0002)/81). It has been also co-funded by the European Union (ERC, InterText, 101054961). Views and opinions expressed are however those of the author(s) only and do not necessarily reflect those of the European Union or the European Research Council. Neither the European Union nor the granting authority can be held responsible for them. 

We gratefully acknowledge the support of Microsoft with a grant for access to OpenAI GPT models via the Azure cloud (Accelerate Foundation Model Academic Research). We thank Furkan Şahinuç, Serwar Basch and Tim Baumgärtner for their valuable feedback and suggestions on a draft of this paper. We would also like to express our gratitude to Kexin Wang, Nils Dycke, Jan Buchmann, Dennis Zyska, Serwar Basch and Falko Helm for their insightful discussions throughout the project.

\bibliography{myref}

\begin{thebibliography}{43}
\expandafter\ifx\csname natexlab\endcsname\relax\def\natexlab#1{#1}\fi

\bibitem[{Chen and Guestrin(2016)}]{XGBoost}
Tianqi Chen and Carlos Guestrin. 2016.
\newblock \href {https://doi.org/10.1145/2939672.2939785} {Xgboost: A scalable tree boosting system}.
\newblock In \emph{Proceedings of the 22nd ACM SIGKDD International Conference on Knowledge Discovery and Data Mining}, KDD '16, page 785–794, New York, NY, USA. Association for Computing Machinery.

\bibitem[{Cortes and Vapnik(1995)}]{svm}
Corinna Cortes and Vladimir Vapnik. 1995.
\newblock \href {https://doi.org/10.1023/A:1022627411411} {Support-vector networks}.
\newblock \emph{Machine Learning}, 20(3):273–297.

\bibitem[{Dettmers et~al.(2023)Dettmers, Pagnoni, Holtzman, and Zettlemoyer}]{qlora}
Tim Dettmers, Artidoro Pagnoni, Ari Holtzman, and Luke Zettlemoyer. 2023.
\newblock \href {http://arxiv.org/abs/2305.14314} {Qlora: Efficient finetuning of quantized llms}.
\newblock \emph{ArXiv}, cs.LG/2305.14314.

\bibitem[{Du et~al.(2022)Du, Raheja, Kumar, Kim, Lopez, and Kang}]{IteraTeR}
Wanyu Du, Vipul Raheja, Dhruv Kumar, Zae~Myung Kim, Melissa Lopez, and Dongyeop Kang. 2022.
\newblock \href {https://doi.org/10.18653/v1/2022.acl-long.250} {Understanding iterative revision from human-written text}.
\newblock In \emph{Proceedings of the 60th Annual Meeting of the Association for Computational Linguistics (Volume 1: Long Papers)}, pages 3573--3590, Dublin, Ireland. Association for Computational Linguistics.

\bibitem[{Dycke et~al.(2023)Dycke, Kuznetsov, and Gurevych}]{nlpeer}
Nils Dycke, Ilia Kuznetsov, and Iryna Gurevych. 2023.
\newblock \href {https://doi.org/10.18653/v1/2023.acl-long.277} {{NLP}eer: A unified resource for the computational study of peer review}.
\newblock In \emph{Proceedings of the 61st Annual Meeting of the Association for Computational Linguistics (Volume 1: Long Papers)}, pages 5049--5073, Toronto, Canada. Association for Computational Linguistics.

\bibitem[{Fortunato et~al.(2018)Fortunato, Bergstrom, Börner, Evans, Helbing, Milojević, Petersen, Radicchi, Sinatra, Uzzi, Vespignani, Waltman, Wang, and Barabási}]{scisci}
Santo Fortunato, Carl~T. Bergstrom, Katy Börner, James~A. Evans, Dirk Helbing, Staša Milojević, Alexander~M. Petersen, Filippo Radicchi, Roberta Sinatra, Brian Uzzi, Alessandro Vespignani, Ludo Waltman, Dashun Wang, and Albert-László Barabási. 2018.
\newblock \href {https://doi.org/10.1126/science.aao0185} {Science of science}.
\newblock \emph{Science}, 359(6379):eaao0185.

\bibitem[{Jiang et~al.(2023)Jiang, Sablayrolles, Mensch, Bamford, Chaplot, de~las Casas, Bressand, Lengyel, Lample, Saulnier, Lavaud, Lachaux, Stock, Scao, Lavril, Wang, Lacroix, and Sayed}]{mistral}
Albert~Q. Jiang, Alexandre Sablayrolles, Arthur Mensch, Chris Bamford, Devendra~Singh Chaplot, Diego de~las Casas, Florian Bressand, Gianna Lengyel, Guillaume Lample, Lucile Saulnier, Lélio~Renard Lavaud, Marie-Anne Lachaux, Pierre Stock, Teven~Le Scao, Thibaut Lavril, Thomas Wang, Timothée Lacroix, and William~El Sayed. 2023.
\newblock \href {http://arxiv.org/abs/2310.06825} {Mistral 7b}.
\newblock \emph{ArXiv}, cs.CL/2310.06825.

\bibitem[{Jiang et~al.(2022)Jiang, Xu, and Stevens}]{arxivedits}
Chao Jiang, Wei Xu, and Samuel Stevens. 2022.
\newblock \href {https://aclanthology.org/2022.emnlp-main.641} {ar{X}iv{E}dits: Understanding the human revision process in scientific writing}.
\newblock In \emph{Proceedings of the 2022 Conference on Empirical Methods in Natural Language Processing}, pages 9420--9435, Abu Dhabi, United Arab Emirates. Association for Computational Linguistics.

\bibitem[{Kashefi et~al.(2022)Kashefi, Afrin, Dale, Olshefski, Godley, Litman, and Hwa}]{argrewrite2}
Omid Kashefi, Tazin Afrin, Meghan Dale, Christopher Olshefski, Amanda Godley, Diane Litman, and Rebecca Hwa. 2022.
\newblock \href {https://doi.org/10.1007/s10579-021-09567-z} {{ArgRewrite} v.2: an annotated argumentative revisions corpus}.
\newblock \emph{Language Resources and Evaluation}, 56(3):881--915.

\bibitem[{Kim et~al.(2024)Kim, Lee, Kwon, Gu, Kim, Cho, yong Sohn, and Choi}]{Linq}
Junseong Kim, Seolhwa Lee, Jihoon Kwon, Sangmo Gu, Yejin Kim, Minkyung Cho, Jy~yong Sohn, and Chanyeol Choi. 2024.
\newblock \href {https://getlinq.com/blog/linq-embed-mistral/} {Linq-embed-mistral:elevating text retrieval with improved gpt data through task-specific control and quality refinement}.
\newblock \emph{Linq AI Research Blog}.

\bibitem[{Kuznetsov et~al.(2022)Kuznetsov, Buchmann, Eichler, and Gurevych}]{f1000rd}
Ilia Kuznetsov, Jan Buchmann, Max Eichler, and Iryna Gurevych. 2022.
\newblock \href {https://doi.org/10.1162/coli_a_00455} {{Revise and Resubmit: An Intertextual Model of Text-based Collaboration in Peer Review}}.
\newblock \emph{Computational Linguistics}, 48(4):949--986.

\bibitem[{Lee et~al.(2024)Lee, Roy, Xu, Raiman, Shoeybi, Catanzaro, and Ping}]{nvembed}
Chankyu Lee, Rajarshi Roy, Mengyao Xu, Jonathan Raiman, Mohammad Shoeybi, Bryan Catanzaro, and Wei Ping. 2024.
\newblock \href {http://arxiv.org/abs/2405.17428} {Nv-embed: Improved techniques for training llms as generalist embedding models}.
\newblock \emph{ArXiv}, cs.CL/2405.17428.

\bibitem[{Lin et~al.(2022)Lin, Hilton, and Evans}]{truthfulqa}
Stephanie Lin, Jacob Hilton, and Owain Evans. 2022.
\newblock \href {https://doi.org/10.18653/v1/2022.acl-long.229} {{T}ruthful{QA}: Measuring how models mimic human falsehoods}.
\newblock In \emph{Proceedings of the 60th Annual Meeting of the Association for Computational Linguistics (Volume 1: Long Papers)}, pages 3214--3252, Dublin, Ireland. Association for Computational Linguistics.

\bibitem[{Liu et~al.(2019)Liu, Ott, Goyal, Du, Joshi, Chen, Levy, Lewis, Zettlemoyer, and Stoyanov}]{roberta}
Yinhan Liu, Myle Ott, Naman Goyal, Jingfei Du, Mandar Joshi, Danqi Chen, Omer Levy, Mike Lewis, Luke Zettlemoyer, and Veselin Stoyanov. 2019.
\newblock \href {https://api.semanticscholar.org/CorpusID:198953378} {Roberta: A robustly optimized bert pretraining approach}.
\newblock \emph{ArXiv}, abs/1907.11692.

\bibitem[{Meng et~al.(2024)Meng, Liu, Joty, Xiong, Zhou, and Yavuz}]{sfr}
Rui Meng, Ye~Liu, Shafiq~Rayhan Joty, Caiming Xiong, Yingbo Zhou, and Semih Yavuz. 2024.
\newblock \href {https://blog.salesforceairesearch.com/sfr-embedded-mistral/} {Sfr-embedding-mistral:enhance text retrieval with transfer learning}.
\newblock \emph{Salesforce AI Research Blog}.

\bibitem[{Milios et~al.(2023)Milios, Reddy, and Bahdanau}]{cls_llm6}
Aristides Milios, Siva Reddy, and Dzmitry Bahdanau. 2023.
\newblock \href {https://doi.org/10.18653/v1/2023.genbench-1.14} {In-context learning for text classification with many labels}.
\newblock In \emph{Proceedings of the 1st GenBench Workshop on (Benchmarking) Generalisation in NLP}, pages 173--184, Singapore. Association for Computational Linguistics.

\bibitem[{Muennighoff et~al.(2023)Muennighoff, Tazi, Magne, and Reimers}]{mteb}
Niklas Muennighoff, Nouamane Tazi, Loic Magne, and Nils Reimers. 2023.
\newblock \href {https://doi.org/10.18653/v1/2023.eacl-main.148} {{MTEB}: Massive text embedding benchmark}.
\newblock In \emph{Proceedings of the 17th Conference of the European Chapter of the Association for Computational Linguistics}, pages 2014--2037, Dubrovnik, Croatia. Association for Computational Linguistics.

\bibitem[{Parikh et~al.(2023)Parikh, Tiwari, Tumbade, and Vohra}]{cls_llm5}
Soham Parikh, Mitul Tiwari, Prashil Tumbade, and Quaizar Vohra. 2023.
\newblock \href {https://doi.org/10.18653/v1/2023.acl-industry.71} {Exploring zero and few-shot techniques for intent classification}.
\newblock In \emph{Proceedings of the 61st Annual Meeting of the Association for Computational Linguistics (Volume 5: Industry Track)}, pages 744--751, Toronto, Canada. Association for Computational Linguistics.

\bibitem[{Patwa et~al.(2024)Patwa, Filice, Chen, Castellucci, Rokhlenko, and Malmasi}]{cls_llm8}
Parth Patwa, Simone Filice, Zhiyu Chen, Giuseppe Castellucci, Oleg Rokhlenko, and Shervin Malmasi. 2024.
\newblock \href {http://arxiv.org/abs/2404.02422} {Enhancing low-resource llms classification with peft and synthetic data}.
\newblock \emph{ArXiv}, cs.CL/2404.02422.

\bibitem[{Peskine et~al.(2023)Peskine, Koren{\v{c}}i{\'c}, Grubisic, Papotti, Troncy, and Rosso}]{cls_llm3}
Youri Peskine, Damir Koren{\v{c}}i{\'c}, Ivan Grubisic, Paolo Papotti, Raphael Troncy, and Paolo Rosso. 2023.
\newblock \href {https://doi.org/10.18653/v1/2023.findings-emnlp.267} {Definitions matter: Guiding {GPT} for multi-label classification}.
\newblock In \emph{Findings of the Association for Computational Linguistics: EMNLP 2023}, pages 4054--4063, Singapore. Association for Computational Linguistics.

\bibitem[{Pham et~al.(2023)Pham, Indurthi, Chollampatt, and Turchi}]{summ_llm2}
Minh-Quang Pham, Sathish Indurthi, Shamil Chollampatt, and Marco Turchi. 2023.
\newblock \href {https://doi.org/10.18653/v1/2023.emnlp-main.753} {Select, prompt, filter: Distilling large language models for summarizing conversations}.
\newblock In \emph{Proceedings of the 2023 Conference on Empirical Methods in Natural Language Processing}, pages 12257--12265, Singapore. Association for Computational Linguistics.

\bibitem[{Qin et~al.(2023)Qin, Zhang, Zhang, Chen, Yasunaga, and Yang}]{cls_llm7}
Chengwei Qin, Aston Zhang, Zhuosheng Zhang, Jiaao Chen, Michihiro Yasunaga, and Diyi Yang. 2023.
\newblock \href {https://doi.org/10.18653/v1/2023.emnlp-main.85} {Is {C}hat{GPT} a general-purpose natural language processing task solver?}
\newblock In \emph{Proceedings of the 2023 Conference on Empirical Methods in Natural Language Processing}, pages 1339--1384, Singapore. Association for Computational Linguistics.

\bibitem[{Raffel et~al.(2020)Raffel, Shazeer, Roberts, Lee, Narang, Matena, Zhou, Li, and Liu}]{t5}
Colin Raffel, Noam Shazeer, Adam Roberts, Katherine Lee, Sharan Narang, Michael Matena, Yanqi Zhou, Wei Li, and Peter~J. Liu. 2020.
\newblock \href {http://jmlr.org/papers/v21/20-074.html} {Exploring the limits of transfer learning with a unified text-to-text transformer}.
\newblock \emph{Journal of Machine Learning Research}, 21(140):1--67.

\bibitem[{Reimers and Gurevych(2019)}]{sbert}
Nils Reimers and Iryna Gurevych. 2019.
\newblock \href {http://arxiv.org/abs/1908.10084} {Sentence-bert: Sentence embeddings using siamese bert-networks}.
\newblock In \emph{Proceedings of the 2019 Conference on Empirical Methods in Natural Language Processing}. Association for Computational Linguistics.

\bibitem[{Rogers and Luccioni(2024)}]{llms_terms_claims}
Anna Rogers and {Alexandra Sasha} Luccioni. 2024.
\newblock \href {https://arxiv.org/abs/2308.07120} {Position: Key claims in llm research have a long tail of footnotes}.
\newblock In \emph{Proceedings of the 41st International Conference on Machine Learning}, volume 235, pages 42647--42665.

\bibitem[{Rouzegar and Makrehchi(2024)}]{cls_llm2}
Hamidreza Rouzegar and Masoud Makrehchi. 2024.
\newblock \href {https://aclanthology.org/2024.law-1.10} {Enhancing text classification through {LLM}-driven active learning and human annotation}.
\newblock In \emph{Proceedings of The 18th Linguistic Annotation Workshop (LAW-XVIII)}, pages 98--111, St. Julians, Malta. Association for Computational Linguistics.

\bibitem[{Ruan et~al.(2024)Ruan, Kuznetsov, and Gurevych}]{re3}
Qian Ruan, Ilia Kuznetsov, and Iryna Gurevych. 2024.
\newblock \href {https://doi.org/10.18653/v1/2024.acl-long.255} {Re3: A holistic framework and dataset for modeling collaborative document revision}.
\newblock In \emph{Proceedings of the 62nd Annual Meeting of the Association for Computational Linguistics (Volume 1: Long Papers)}, pages 4635--4655, Bangkok, Thailand. Association for Computational Linguistics.

\bibitem[{Saravia et~al.(2018)Saravia, Liu, Huang, Wu, and Chen}]{mteb_ec}
Elvis Saravia, Hsien-Chi~Toby Liu, Yen-Hao Huang, Junlin Wu, and Yi-Shin Chen. 2018.
\newblock \href {https://doi.org/10.18653/v1/D18-1404} {{CARER}: Contextualized affect representations for emotion recognition}.
\newblock In \emph{Proceedings of the 2018 Conference on Empirical Methods in Natural Language Processing}, pages 3687--3697, Brussels, Belgium. Association for Computational Linguistics.

\bibitem[{Shah et~al.(2018)Shah, Lei, Moschitti, Romeo, and Nakov}]{mteb_SDQ}
Darsh Shah, Tao Lei, Alessandro Moschitti, Salvatore Romeo, and Preslav Nakov. 2018.
\newblock \href {https://doi.org/10.18653/v1/D18-1131} {Adversarial domain adaptation for duplicate question detection}.
\newblock In \emph{Proceedings of the 2018 Conference on Empirical Methods in Natural Language Processing}, pages 1056--1063, Brussels, Belgium. Association for Computational Linguistics.

\bibitem[{Student(1908)}]{ttest}
Student. 1908.
\newblock \href {http://www.jstor.org/stable/2331554} {The probable error of a mean}.
\newblock \emph{Biometrika}, 6(1):1--25.

\bibitem[{Su and Yan(2017)}]{mteb_tuc}
Yu~Su and Xifeng Yan. 2017.
\newblock \href {https://doi.org/10.18653/v1/D17-1127} {Cross-domain semantic parsing via paraphrasing}.
\newblock In \emph{Proceedings of the 2017 Conference on Empirical Methods in Natural Language Processing}, pages 1235--1246, Copenhagen, Denmark. Association for Computational Linguistics.

\bibitem[{Sun et~al.(2023)Sun, Li, Li, Wu, Guo, Zhang, and Wang}]{cls_llm}
Xiaofei Sun, Xiaoya Li, Jiwei Li, Fei Wu, Shangwei Guo, Tianwei Zhang, and Guoyin Wang. 2023.
\newblock \href {https://doi.org/10.18653/v1/2023.findings-emnlp.603} {Text classification via large language models}.
\newblock In \emph{Findings of the Association for Computational Linguistics: EMNLP 2023}, pages 8990--9005, Singapore. Association for Computational Linguistics.

\bibitem[{Touvron et~al.(2023)Touvron, Martin, Stone, Albert, Almahairi, Babaei, Bashlykov, Batra, Bhargava, Bhosale, Bikel, Blecher, Ferrer, Chen, Cucurull, Esiobu, Fernandes, Fu, Fu, Fuller, Gao, Goswami, Goyal, Hartshorn, Hosseini, Hou, Inan, Kardas, Kerkez, Khabsa, Kloumann, Korenev, Koura, Lachaux, Lavril, Lee, Liskovich, Lu, Mao, Martinet, Mihaylov, Mishra, Molybog, Nie, Poulton, Reizenstein, Rungta, Saladi, Schelten, Silva, Smith, Subramanian, Tan, Tang, Taylor, Williams, Kuan, Xu, Yan, Zarov, Zhang, Fan, Kambadur, Narang, Rodriguez, Stojnic, Edunov, and Scialom}]{llama2}
Hugo Touvron, Louis Martin, Kevin~R. Stone, Peter Albert, Amjad Almahairi, Yasmine Babaei, Nikolay Bashlykov, Soumya Batra, Prajjwal Bhargava, Shruti Bhosale, D.~Bikel, Lukas Blecher, Cristian~Cantón Ferrer, Moya Chen, Guillem Cucurull, David Esiobu, Jude Fernandes, Jeremy Fu, Wenyin Fu, Brian Fuller, Cynthia Gao, Vedanuj Goswami, Naman Goyal, A.~Hartshorn, Saghar Hosseini, Rui Hou, Hakan Inan, Marcin Kardas, Viktor Kerkez, Madian Khabsa, Isabel~M. Kloumann, A.~Korenev, Punit~Singh Koura, Marie-Anne Lachaux, Thibaut Lavril, Jenya Lee, Diana Liskovich, Yinghai Lu, Yuning Mao, Xavier Martinet, Todor Mihaylov, Pushkar Mishra, Igor Molybog, Yixin Nie, Andrew Poulton, Jeremy Reizenstein, Rashi Rungta, Kalyan Saladi, Alan Schelten, Ruan Silva, Eric~Michael Smith, R.~Subramanian, Xia Tan, Binh Tang, Ross Taylor, Adina Williams, Jian~Xiang Kuan, Puxin Xu, Zhengxu Yan, Iliyan Zarov, Yuchen Zhang, Angela Fan, Melanie Kambadur, Sharan Narang, Aurelien Rodriguez, Robert Stojnic, Sergey Edunov, and Thomas Scialom. 2023.
\newblock \href {https://doi.org/10.48550/arXiv.2307.09288} {Llama 2: Open foundation and fine-tuned chat models}.
\newblock \emph{ArXiv}, abs/2307.09288.

\bibitem[{Tsoumakas et~al.(2011)Tsoumakas, Spyromitros-Xioufis, Vilcek, and Vlahavas}]{mulan}
Grigorios Tsoumakas, Eleftherios Spyromitros-Xioufis, Jozef Vilcek, and Ioannis Vlahavas. 2011.
\newblock \href {http://jmlr.org/papers/v12/tsoumakas11a.html} {Mulan: A java library for multi-label learning}.
\newblock \emph{Journal of Machine Learning Research}, 12(71):2411--2414.

\bibitem[{Wang and Komatsuzaki(2021)}]{gpt-j}
Ben Wang and Aran Komatsuzaki. 2021.
\newblock {GPT-J-6B: A 6 Billion Parameter Autoregressive Language Model}.
\newblock \url{https://github.com/kingoflolz/mesh-transformer-jax}.

\bibitem[{Wang et~al.(2023)Wang, Zhang, and Wang}]{summ_llm}
Yiming Wang, Zhuosheng Zhang, and Rui Wang. 2023.
\newblock \href {https://doi.org/10.18653/v1/2023.acl-long.482} {Element-aware summarization with large language models: Expert-aligned evaluation and chain-of-thought method}.
\newblock In \emph{Proceedings of the 61st Annual Meeting of the Association for Computational Linguistics (Volume 1: Long Papers)}, pages 8640--8665, Toronto, Canada. Association for Computational Linguistics.

\bibitem[{Wilcoxon(1945)}]{wilcoxon}
Frank Wilcoxon. 1945.
\newblock \href {http://www.jstor.org/stable/3001968} {Individual comparisons by ranking methods}.
\newblock \emph{Biometrics Bulletin}, 1(6):80--83.

\bibitem[{Xu et~al.(2015)Xu, Callison-Burch, and Dolan}]{mteb_tse}
Wei Xu, Chris Callison-Burch, and Bill Dolan. 2015.
\newblock \href {https://doi.org/10.18653/v1/S15-2001} {{S}em{E}val-2015 task 1: Paraphrase and semantic similarity in {T}witter ({PIT})}.
\newblock In \emph{Proceedings of the 9th International Workshop on Semantic Evaluation ({S}em{E}val 2015)}, pages 1--11, Denver, Colorado. Association for Computational Linguistics.

\bibitem[{Yang et~al.(2017)Yang, Halfaker, Kraut, and Hovy}]{wikieditintent-yang}
Diyi Yang, Aaron Halfaker, Robert Kraut, and Eduard Hovy. 2017.
\newblock \href {https://doi.org/10.18653/v1/D17-1213} {Identifying semantic edit intentions from revisions in {W}ikipedia}.
\newblock In \emph{Proceedings of the 2017 Conference on Empirical Methods in Natural Language Processing}, pages 2000--2010, Copenhagen, Denmark. Association for Computational Linguistics.

\bibitem[{Zellers et~al.(2019)Zellers, Holtzman, Bisk, Farhadi, and Choi}]{hellaswag}
Rowan Zellers, Ari Holtzman, Yonatan Bisk, Ali Farhadi, and Yejin Choi. 2019.
\newblock \href {https://doi.org/10.18653/v1/P19-1472} {{H}ella{S}wag: Can a machine really finish your sentence?}
\newblock In \emph{Proceedings of the 57th Annual Meeting of the Association for Computational Linguistics}, pages 4791--4800, Florence, Italy. Association for Computational Linguistics.

\bibitem[{Zhang et~al.(2016)Zhang, Hwa, Litman, and Hashemi}]{argrewrite}
Fan Zhang, Rebecca Hwa, Diane Litman, and Homa~B. Hashemi. 2016.
\newblock \href {https://doi.org/10.18653/v1/N16-3008} {{A}rg{R}ewrite: A web-based revision assistant for argumentative writings}.
\newblock In \emph{Proceedings of the 2016 Conference of the North {A}merican Chapter of the Association for Computational Linguistics: Demonstrations}, pages 37--41, San Diego, California. Association for Computational Linguistics.

\bibitem[{Zhang et~al.(2023)Zhang, Khalifa, Logeswaran, Lee, Lee, and Wang}]{qa_llm2}
Yunxiang Zhang, Muhammad Khalifa, Lajanugen Logeswaran, Moontae Lee, Honglak Lee, and Lu~Wang. 2023.
\newblock \href {https://doi.org/10.18653/v1/2023.emnlp-main.286} {Merging generated and retrieved knowledge for open-domain {QA}}.
\newblock In \emph{Proceedings of the 2023 Conference on Empirical Methods in Natural Language Processing}, pages 4710--4728, Singapore. Association for Computational Linguistics.

\bibitem[{Zhong and Chen(2021)}]{PURE}
Zexuan Zhong and Danqi Chen. 2021.
\newblock \href {https://doi.org/10.18653/v1/2021.naacl-main.5} {A frustratingly easy approach for entity and relation extraction}.
\newblock In \emph{Proceedings of the 2021 Conference of the North American Chapter of the Association for Computational Linguistics: Human Language Technologies}, pages 50--61, Online. Association for Computational Linguistics.

\end{thebibliography}
\bibliographystyle{acl_natbib}

\appendix

\section{Framework}
\label{sec:method_app}

\noindent \textbf{Input Tuning.} Table \ref{tab:input_examples} provides examples of input texts in various settings, see §\ref{subsec:input_tuning} for details on input tuning. 

\noindent \textbf{Language Models.} 
We select the LLMs based on four criteria: (1) they should be open-sourced to ensure reproducibility; (2) they should have a reasonable size to allow fine-tuning using QLoRA \cite{qlora} with moderate computing resources, while still varying in size (ranging from 6B to 13B) to assess the impact of model size; (3) there should be both instruction-fine-tuned and non-instruction-fine-tuned versions to study their performance differences and evaluate the effectiveness of instruction fine-tuning for different approaches (see RQ2 in §\ref{subsec:diss}), and (4)  they should be recent and proven to be state-of-the-art or advanced on extensive NLP benchmarks \cite{hellaswag, truthfulqa, mteb}.\footnote{As of April 2024}
We select the small pre-trained language models (PLMs) that can be fully fine-tuned with equivalent computing resources. 
For the generation-based approach, we select an encoder-decoder PLM (i.e., T5) specifically designed for text-to-text generation to align with the approach's design. For the encoding-based approach, we use an encoder-only transformer model (i.e., RoBERTa) to assess its encoding capabilities in comparison to LLMs.
Table \ref{tab:model_comp} compares the models' features, including parameter size, number of layers, model dimension and architecture.

\begin{table}[t]
\fontsize{8}{8}
\selectfont
\centering
\renewcommand{\arraystretch}{1.4} 
\tabcolsep=0.07cm
\begin{tabular}[]{llllll}
\toprule
models&size & \#layers & dim & inst&architecture\\\hline
GPT-j \citeyearpar{gpt-j} &6B &28&4096&no&decoder-only\\
Mistral-Instruct \citeyearpar{mistral}&7B &32&4096&yes&decoder-only\\
Llama2-7B \citeyearpar{llama2}&7B &32&4096&no&decoder-only\\
Llama2-7B-Chat \citeyearpar{llama2}&7B &32&4096&yes&decoder-only\\
Llama2-13B \citeyearpar{llama2}&13B &40&5120&no&decoder-only\\
Llama2-13B-Chat \citeyearpar{llama2}&13B &40&5120&yes&decoder-only\\
Llama3-8B (2024)&8B &32&4096&no&decoder-only\\
Llama3-8B-Instruct (2024)&8B &32&4096&yes&decoder-only\\\hline
RoBERTa-base \citeyearpar{roberta}&125M&12&768&no&encoder-only\\
T5-base \citeyearpar{t5}&220M&12&768&no&encoder-decoder\\
\bottomrule
\end{tabular}
\caption[]{Language model comparisons. Presented are the parameter size, number of layers, model dimension, whether the model is fine-tuned for instruction-following, and the transformer architecture of each model. }
\label{tab:model_comp}
\end{table}

\section{Experimental Details}
\label{sec:exp_details}
We fine-tune all linear layers of the LLMs using QLoRA \cite{qlora}, tuning parameters such as LoRA rank ($r$), LoRA alpha ($a$), and dropout ($d$) during initial experiments.
Based on the results in Table \ref{tab:hyper_tune}, we set the parameters as follows: for approach \textit{Gen}, 
we set $r$=256, $a$= 256, $d$=0.1; for approaches \textit{SeqC}, \textit{SNet}, and \textit{XNet}, the settings are $r$=128, $a$=128, $d$=0.1. The small PLMs, T5 and RoBERTa, are fully fine-tuned with all weights being directly updated. 

\begin{table}[t]
\fontsize{8}{8}
\selectfont
\centering
\renewcommand{\arraystretch}{1.17} 
\begin{tabular}[]{lllllll} \toprule
base LM & r & a&d&acc.&m.f1&AIR\\ \hline
\multicolumn{7}{c}{\textbf{(a). Gen}} \\ \hline
Llama2-13B-Chat&16&16&0.1&81.5&80.7&100\\
&128&16&0.1&81.8&81.1&100\\
&128&128&0.1&82.4&81.5&100\\
&256&16&0.1&80.8&80.7&100\\
&256&128&0.1&83.1&81.9&99.9\\
&256&256&0.1&\textbf{83.6}&\textbf{82.8}&100\\
&256&512&0.1&79.5&79.3&94.1\\
&512&16&0.1&81.7&80.3&99.9\\
&512&512&0.1&82.3&80.9&99.8\\
\hline
\multicolumn{7}{c}{\textbf{(b). SeqC}} \\ \hline
Llama2-7B-Chat&16&16&0.1&83.9&82.2&100\\
&64&64&0.1&83.7&82.3&100\\
&128&128&0.1&\textbf{84.4}&\textbf{82.8}&100\\
&128&128&0.2&84.1&82.5&100\\
&256&256&0.1&83.8&82.0&100\\
&512&512&0.1&81.7&80.5&100\\
\bottomrule
\end{tabular}
\caption[LLMs]{Hyperparameters tuning. r: LoRA rank, a: LORA alpha, d: dropout. acc.: accuracy, m.f1: marco F1 score, AIR: Answer Inclusion Rate.}
\label{tab:hyper_tune}
\end{table}

\begin{table*}[ht]
\fontsize{9}{9}
\selectfont
\renewcommand{\arraystretch}{1.2} 
\tabcolsep=0.13cm
\begin{tabular}{lll}
\toprule
\textbf{(a) Gen} & &
{\textcircled{\tiny 1} \textit{ inst + natural input}} \\\hline
&\multicolumn{2}{l}{\begin{tabular}{l}
Instruction: Classify the intent of the following sentence edit. The possible labels  are: Grammar, Clarity, \\Fact/Evidence, Claim, Other.  \\ INPUT:\\ OLD: The model is trained in a NVIDIA GeForce RTX 2080Ti GPU. \\ NEW: The model is trained in an NVIDIA GeForce RTX 2080Ti GPU.\\RESPONSE:
\end{tabular}}\\ \cline{2-3}
&&{\textcircled{\tiny 2} \textit{ inst + structured input}}  \\\cline{2-3}
&\multicolumn{2}{l}{\begin{tabular}{l}
<instruction>\\ Classify the intent of the following sentence edit. The possible labels are: Grammar, Clarity, \\Fact/Evidence, Claim, Other.  \\</instruction>\\<input>\\ <old> The model is trained in a NVIDIA GeForce RTX 2080Ti GPU. </old>\\ <new> The model is trained in an NVIDIA GeForce RTX 2080Ti GPU. </new>\\</input>\\<response>
\end{tabular}}\\  \hline
\textbf{(b) SeqC} & &
{\textcircled{\tiny 1} \textit{natural input}}   \\\hline
&\multicolumn{2}{l}{\begin{tabular}{l}
The model is trained in a NVIDIA GeForce RTX 2080Ti GPU.\\
The model is trained in an NVIDIA GeForce RTX 2080Ti GPU.
\end{tabular}}\\ \cline{2-3}
&&{\textcircled{\tiny 2} \textit{structured input}} \\\cline{2-3}
&\multicolumn{2}{l}{\begin{tabular}{l}
<old> The model is trained in a NVIDIA GeForce RTX 2080Ti GPU. </old>\\ <new> The model is trained in an NVIDIA GeForce RTX 2080Ti GPU. </new>
\end{tabular}}\\ \cline{2-3}
&&{\textcircled{\tiny 3} \textit{inst + structured input}}   \\\cline{2-3}
&\multicolumn{2}{l}{\begin{tabular}{l}
Classify the intent of the following sentence edit. The possible labels are: Grammar, Clarity, \\ Fact/Evidence, Claim, Other. \\
<old> The model is trained in a NVIDIA GeForce RTX 2080Ti GPU. </old>\\ <new> The model is trained in an NVIDIA GeForce RTX 2080Ti GPU. </new>
\end{tabular}}\\ 
\bottomrule
\end{tabular}
\caption{Examples of different input types.}
\label{tab:input_examples}
\end{table*}

For approach \textit{Gen}, the output token limit is set to ten. We define the metric \textit{Answer Inclusion Rate} (\textit{AIR}) as the percentage of samples where a label string falls within the ten output tokens regardless of correctness. If the output tokens do not contain any label string, the prediction is considered a failure. When using RoBERTa for approach \textit{SeqC}, the the first token representation is used as the input for classification.

For all approaches and base LMs, the models are fine-tuned for ten epochs on the training set, with checkpoints saved after each epoch. The final model selection is determined based on evaluation results from the validation set, and its performance is subsequently assessed on the test set. For approaches \textit{SeqC}, \textit{SNet}, and \textit{XNet}, a single NVIDIA A100 or H100 GPU with 80GB memory is utilized. Approach \textit{Gen} requires two such GPUs.

In Table \ref{tab:app_a_b}, the human performance is calculated from individual human annotations in Re3-Sci and the gold labels aggregated by majority voting. For the GPT-4 baselines, the gpt-4-turbo model released in April 2024 was used. GPT-4 (ICL+CoT) uses the default ICL examples and CoT formats provided by \citet{re3}. In Table \ref{tab:app_c_d}, the structured input format (§\ref{subsec:input_tuning}) without task instructions is used.

\begin{table*}[t]
\fontsize{8}{8}
\selectfont
\centering
\renewcommand{\arraystretch}{1.3} 
\tabcolsep=0.06cm
\begin{tabular}[]{ll|lll|lll|llll|lll|lll}
\hline
Task Type&&\multicolumn{9}{c}{Binary Pair Classification}
&&\multicolumn{6}{c}{Single Input Classification}\\ \hline
Dataset&&&SDQ&&&TSE&&&TUC&&&&EC&&&TSEC\\ \hline
Metric&& acc. & m.f1 & AIR & acc. & m.f1 & AIR & acc. & m.f1 & AIR 
&& acc. & m.f1 & AIR & acc. & m.f1 & AIR \\ \hline
Previous SOTA
&&\begin{tabular}{l}99.9\end{tabular}&\begin{tabular}{l}93.7\end{tabular}
&&\begin{tabular}{l}88.4\end{tabular}&\begin{tabular}{l}73.7\end{tabular}
&&\begin{tabular}{l}90.0\end{tabular}&\begin{tabular}{l}80.5\end{tabular}
&&&\begin{tabular}{l}93.4\end{tabular}&\begin{tabular}{l}90.1\end{tabular}
&&\begin{tabular}{l}80.9\end{tabular}&\begin{tabular}{l}81.2\end{tabular}\\\hline
T5
&\begin{tabular}{l}\textit{Gen}\end{tabular} 
&\begin{tabular}{l}99.1\end{tabular}
&\begin{tabular}{l}86.4\end{tabular}
&\begin{tabular}{l}100\end{tabular}
&\begin{tabular}{l}87.9\end{tabular}
&\begin{tabular}{l}83.8\end{tabular}
&\begin{tabular}{l}100\end{tabular}
&\begin{tabular}{l}86.8\end{tabular}
&\begin{tabular}{l}84.1\end{tabular}
&\begin{tabular}{l}100\end{tabular}

&&\begin{tabular}{l}93.1\end{tabular}
&\begin{tabular}{l}89.5\end{tabular}
&\begin{tabular}{l}100\end{tabular}
&\begin{tabular}{l}80.0\end{tabular}
&\begin{tabular}{l}80.1\end{tabular}
&\begin{tabular}{l}100\end{tabular}
\\ \hline
RoBERTa 
&\begin{tabular}{l}\textit{SeqC}\end{tabular} 
&\begin{tabular}{l}99.7\end{tabular}
&\begin{tabular}{l}93.5\end{tabular}
&\begin{tabular}{l}100\end{tabular}
&\begin{tabular}{l}88.6\end{tabular}
&\begin{tabular}{l}79.4\end{tabular}
&\begin{tabular}{l}100\end{tabular}
&\begin{tabular}{l}88.5\end{tabular}
&\begin{tabular}{l}85.9\end{tabular}
&\begin{tabular}{l}100\end{tabular}
&&\begin{tabular}{l}93.2\end{tabular}
&\begin{tabular}{l}89.3\end{tabular}
&\begin{tabular}{l}100\end{tabular}
&\begin{tabular}{l}79.1\end{tabular}
&\begin{tabular}{l}79.3\end{tabular}
&\begin{tabular}{l}100\end{tabular}
\\ \hline
GPT-j 
&\begin{tabular}{l}\textit{Gen}\\\textit{SeqC}\\\textit{SNet}\\\textit{XNet}\end{tabular}
&\begin{tabular}{l}96.2\\\underline{99.8}\\98.7\\98.7\end{tabular}
&\begin{tabular}{l}69.2\\\underline{95.6}\\49.7\\49.7\end{tabular}
&\begin{tabular}{l}100\\100\\100\\100\end{tabular}
&\begin{tabular}{l}81.4\\\underline{89.4}\\78.1\\78.1\end{tabular}
&\begin{tabular}{l}76.6\\\underline{84.8}\\43.8\\43.8\end{tabular}
&\begin{tabular}{l}99.7\\100\\100\\100\end{tabular}
&\begin{tabular}{l}86.2\\\underline{90.2}\\73.9\\73.9\end{tabular}
&\begin{tabular}{l}82.9\\\underline{87.5}\\42.5\\42.5\end{tabular}
&\begin{tabular}{l}99.9\\100\\100\\100\end{tabular}
&&\begin{tabular}{l}90.0\\\underline{93.0}\\-\\-\end{tabular}
&\begin{tabular}{l}85.2\\\underline{88.2}\\-\\-\end{tabular}
&\begin{tabular}{l}100\\100\\-\\-\end{tabular}
&\begin{tabular}{l}65.9\\\underline{78.3}\\-\\-\end{tabular}
&\begin{tabular}{l}69.1\\\underline{78.6}\\-\\-\end{tabular}
&\begin{tabular}{l}91.4\\100\\-\\-\end{tabular}
\\ \hline
Mistral-Instruct
&\begin{tabular}{l}\textit{Gen}\\\textit{SeqC}\\\textit{SNet}\\\textit{XNet}\end{tabular}
&\begin{tabular}{l}99.8\\\underline{99.9}\\98.2\\99.7\end{tabular}
&\begin{tabular}{l}95.7\\\underline{97.3}\\79.1\\94.2\end{tabular}
&\begin{tabular}{l}100\\100\\100\\100\end{tabular}

&\begin{tabular}{l}89.7\\78.1\\68.5\\\underline{90.9}\textsuperscript{\#}\end{tabular}
&\begin{tabular}{l}85.5\\43.8\\55.5\\\underline{86.6}\end{tabular}
&\begin{tabular}{l}100\\100\\100\\100\end{tabular}

&\begin{tabular}{l}75.5\\\underline{90.5}\\83.0\\87.2\end{tabular}
&\begin{tabular}{l}50.3\\\underline{87.8}\\76.5\\84.6\end{tabular}
&\begin{tabular}{l}100\\100\\100\\100\end{tabular}

&&\begin{tabular}{l}90.5\\\underline{92.5}\\-\\-\end{tabular}
&\begin{tabular}{l}84.9\\\underline{88.2}\\-\\-\end{tabular}
&\begin{tabular}{l}99.9\\100\\-\\-\end{tabular}

&\begin{tabular}{l}\underline{80.0}\textsuperscript{$\dagger$}\\79.2\\-\\-\end{tabular}
&\begin{tabular}{l}\underline{80.3}\textsuperscript{$\dagger$}\\79.5\\-\\-\end{tabular}
&\begin{tabular}{l}100\\100\\-\\-\end{tabular}
\\ \hline
Llama2-7B  
&\begin{tabular}{l}\textit{Gen}\\\textit{SeqC}\\\textit{SNet}\\\textit{XNet}\end{tabular}
&\begin{tabular}{l}99.5\\99.6\\99.3\\\underline{99.7}\end{tabular}
&\begin{tabular}{l}92.1\\93.4\\89.2\\\underline{94.2}\end{tabular}
&\begin{tabular}{l}100\\100\\100\\100\end{tabular}

&\begin{tabular}{l}89.5\\89.6\\89.6\\\underline{90.8}\end{tabular}
&\begin{tabular}{l}84.5\\86.0\\84.9\\\underline{87.2}\textsuperscript{\#}\end{tabular}
&\begin{tabular}{l}100\\100\\100\\100\end{tabular}

&\begin{tabular}{l}\underline{90.1}\\90.0\\88.8\textsuperscript{!}\\89.4\end{tabular}
&\begin{tabular}{l}87.6\\\underline{87.8}\\85.4\\86.6\end{tabular}
&\begin{tabular}{l}100\\100\\100\\100\end{tabular}

&&\begin{tabular}{l}79.8\\\underline{93.5}\\-\\-\end{tabular}
&\begin{tabular}{l}71.6\\\underline{89.2}\\-\\-\end{tabular}
&\begin{tabular}{l}100\\100\\-\\-\end{tabular}

&\begin{tabular}{l}73.0\\\underline{77.8}\\-\\-\end{tabular}
&\begin{tabular}{l}73.9\\\underline{78.2}\\-\\-\end{tabular}
&\begin{tabular}{l}98.7\\100\\-\\-\end{tabular}
\\ \hline
Llama2-7B-Chat 
&\begin{tabular}{l}\textit{Gen}\\\textit{SeqC}\\\textit{SNet}\\\textit{XNet}\end{tabular}
&\begin{tabular}{l}99.5\\\underline{99.9}\\99.5\\99.4\end{tabular}
&\begin{tabular}{l}91.5\\\underline{98.2}\\91.7\\90.8\end{tabular}
&\begin{tabular}{l}100\\100\\100\\100\end{tabular}

&\begin{tabular}{l}88.6\\89.9\\87.1\\\underline{90.4}\end{tabular}
&\begin{tabular}{l}84.9\\85.9\\81.3\\\underline{86.5}\end{tabular}
&\begin{tabular}{l}100\\100\\100\\100\end{tabular}

&\begin{tabular}{l}89.9\\\underline{90.9}\\88.4\\90.7\end{tabular}
&\begin{tabular}{l}87.3\\\underline{88.5}\\84.0\\88.3\end{tabular}
&\begin{tabular}{l}100\\100\\100\\100\end{tabular}

&&\begin{tabular}{l}89.3\\\underline{93.8}\\-\\-\end{tabular}
&\begin{tabular}{l}84.5\\\underline{\textbf{89.6}}\textsuperscript{*}\\-\\-\end{tabular}
&\begin{tabular}{l}99.7\\100\\-\\-\end{tabular}

&\begin{tabular}{l}75.5\\\underline{78.3}\\-\\-\end{tabular}
&\begin{tabular}{l}75.8\\\underline{78.6}\\-\\-\end{tabular}
&\begin{tabular}{l}100\\100\\-\\-\end{tabular}
\\ \hline
Llama2-13B 
&\begin{tabular}{l}\textit{Gen}\\\textit{SeqC}\\\textit{SNet}\\\textit{XNet}\end{tabular}
&\begin{tabular}{l}99.6\\99.7\\99.8\\\underline{99.9}\end{tabular}
&\begin{tabular}{l}92.7\\94.9\\96.4\\\underline{98.2}\end{tabular}
&\begin{tabular}{l}100\\100\\100\\100\end{tabular}
&\begin{tabular}{l}89.3\\\underline{\textbf{91.3}}\textsuperscript{*}\\88.4\\89.4\end{tabular}
&\begin{tabular}{l}85.2\\\underline{\textbf{87.6}}\textsuperscript{*}\\84.5\\83.5\end{tabular}
&\begin{tabular}{l}100\\100\\100\\100\end{tabular}

&\begin{tabular}{l}\underline{90.9}\\\underline{90.9}\\88.2\\90.1\end{tabular}
&\begin{tabular}{l}\underline{88.6}\textsuperscript{$\dagger$}\\88.2\\85.4\\86.5\end{tabular}
&\begin{tabular}{l}100\\100\\100\\100\end{tabular}

&&\begin{tabular}{l}90.0\\\underline{93.2}\\-\\-\end{tabular}
&\begin{tabular}{l}84.8\\\underline{88.8}\\-\\-\end{tabular}
&\begin{tabular}{l}99.9\\100\\-\\-\end{tabular}

&\begin{tabular}{l}76.3\\\underline{79.0}\\-\\-\end{tabular}
&\begin{tabular}{l}76.8\\\underline{79.4}\\-\\-\end{tabular}
&\begin{tabular}{l}100\\100\\-\\-\end{tabular}
\\ \hline
Llama2-13B-Chat 
&\begin{tabular}{l}\textit{Gen}\\\textit{SeqC}\\\textit{SNet}\\\textit{XNet}\end{tabular}
&\begin{tabular}{l}\underline{99.9}\textsuperscript{$\dagger$}\\99.7\\99.6\\99.8\end{tabular}
&\begin{tabular}{l}\underline{98.2}\textsuperscript{$\dagger$}\\94.2\\92.5\\95.7\end{tabular}
&\begin{tabular}{l}100\\100\\100\\100\end{tabular}

&\begin{tabular}{l}89.9\textsuperscript{$\dagger$}\\\underline{90.5}\\90.4\textsuperscript{!}\\89.9\end{tabular}
&\begin{tabular}{l}86.4\textsuperscript{$\dagger$}\\\underline{86.8}\\85.7\textsuperscript{!}\\86.4\end{tabular}
&\begin{tabular}{l}100\\100\\100\\100\end{tabular}

&\begin{tabular}{l}91.0\textsuperscript{$\dagger$}\\\underline{\textbf{91.3}}\textsuperscript{*}\\88.7\\91.0\textsuperscript{\#}\end{tabular}
&\begin{tabular}{l}88.2\\\underline{\textbf{88.9}}\textsuperscript{*}\\85.5\\88.6\textsuperscript{\#}\end{tabular}
&\begin{tabular}{l}100\\100\\100\\100\end{tabular}

&&\begin{tabular}{l}91.0\textsuperscript{$\dagger$}\\\underline{93.1}\\-\\-\end{tabular}
&\begin{tabular}{l}86.9\textsuperscript{$\dagger$}\\\underline{87.7}\\-\\-\end{tabular}
&\begin{tabular}{l}99.9\\100\\-\\-\end{tabular}

&\begin{tabular}{l}77.0\\\underline{79.8}\\-\\-\end{tabular}
&\begin{tabular}{l}77.7\\\underline{80.1}\\-\\-\end{tabular}
&\begin{tabular}{l}98.5\\100\\-\\-\end{tabular}
\\ \hline
Llama3-8B 
&\begin{tabular}{l}\textit{Gen}\\\textit{SeqC}\\\textit{SNet}\\\textit{XNet}\end{tabular}
&\begin{tabular}{l}99.6\\99.7\\99.2\\\underline{99.8}\end{tabular}
&\begin{tabular}{l}92.2\\94.0\\87.5\\\underline{95.7}\end{tabular}
&\begin{tabular}{l}100\\100\\100\\100\end{tabular}

&\begin{tabular}{l}89.9\textsuperscript{$\dagger$}\\\underline{91.0}\\89.0\\90.2\end{tabular}
&\begin{tabular}{l}86.4\textsuperscript{$\dagger$}\\\underline{87.1}\\84.4\\87.0\end{tabular}
&\begin{tabular}{l}100\\100\\100\\100\end{tabular}

&\begin{tabular}{l}89.7\\\underline{89.8}\\86.7\\\underline{89.8}\end{tabular}
&\begin{tabular}{l}87.1\\\underline{87.2}\\80.5\\86.7\end{tabular}
&\begin{tabular}{l}100\\100\\100\\100\end{tabular}

&&\begin{tabular}{l}56.5\\\underline{93.5}\\-\\-\end{tabular}
&\begin{tabular}{l}47.1\\\underline{88.7}\\-\\-\end{tabular}
&\begin{tabular}{l}98.5\\100\\-\\-\end{tabular}

&\begin{tabular}{l}75.6\\\underline{\textbf{80.5}}\textsuperscript{*}\\-\\-\end{tabular}
&\begin{tabular}{l}75.9\\\underline{\textbf{80.8}}\textsuperscript{*}\\-\\-\end{tabular}
&\begin{tabular}{l}100\\100\\-\\-\end{tabular}
\\ \hline
Llama3-8B-Instruct 
&\begin{tabular}{l}\textit{Gen}\\\textit{SeqC}\\\textit{SNet}\\\textit{XNet}\end{tabular}
&\begin{tabular}{l}99.3\\\underline{\textbf{100}}\textsuperscript{*}\\99.9\textsuperscript{!}\\\underline{\textbf{100}}\textsuperscript{\#}\end{tabular}
&\begin{tabular}{l}88.9\\\underline{\textbf{99.1}}\textsuperscript{*}\\97.2\textsuperscript{!}\\\underline{\textbf{99.1}}\textsuperscript{\#}\end{tabular}
&\begin{tabular}{l}100\\100\\100\\100\end{tabular}

&\begin{tabular}{l}89.5\\\underline{90.6}\\88.1\\89.6\end{tabular}
&\begin{tabular}{l}86.1\\\underline{87.0}\\82.9\\86.3\end{tabular}
&\begin{tabular}{l}100\\100\\100\\100\end{tabular}

&\begin{tabular}{l}88.6\\89.8\\88.6\\\underline{90.4}\end{tabular}
&\begin{tabular}{l}84.3\\87.4\\85.6\textsuperscript{!}\\\underline{88.0}\end{tabular}
&\begin{tabular}{l}100\\100\\100\\100\end{tabular}

&&\begin{tabular}{l}56.7\\\underline{\textbf{94.1}}\textsuperscript{*}\\-\\-\end{tabular}
&\begin{tabular}{l}47.7\\\underline{\textbf{89.6}}\textsuperscript{*}\\-\\-\end{tabular}
&\begin{tabular}{l}99.7\\100\\-\\-\end{tabular}

&\begin{tabular}{l}77.4\\\underline{78.6}\\-\\-\end{tabular}
&\begin{tabular}{l}77.8\\\underline{78.9}\\-\\-\end{tabular}
&\begin{tabular}{l}100\\100\\-\\-\end{tabular}
\\ \hline
\end{tabular}
\caption[LLMs]{Results on the MTEB benchmark. Reported are accuracy (acc.), macro average F1 score (m. f1) and Answer Inclusion Rate (AIR) on the test set of each dataset. The best metrics for each dataset are in bold. The best-performing approach for each LM is underlined. \textsuperscript{$\dagger$}, \textsuperscript{*}, \textsuperscript{!} and \textsuperscript{\#} denote the best-performing LM within each approach (Gen, SeqC, SNet and XNet) for each dataset.  }
\label{tab:gen}
\end{table*}

\section{Generalization Evaluation}
\label{sec:gen}
We assess the generalization of our findings from the EIC dataset across five additional classification tasks from the MTEB benchmark \cite{mteb}.
\subsection{Tasks and Datasets}
\label{subsec:gen_datasets}
The selected tasks comprise three binary pair classification tasks, in which the inputs are sentence pairs and the outputs are binary labels, along with two multi-class single-input classification tasks, where the inputs consist of individual sentences and the output labels are multi-class. The former group features a similar input architecture to that of EIC, allowing for the application of all four approaches. The latter type exhibits output complexity comparable to that of EIC, encompassing multiple potential labels. Below, we provide brief descriptions of each dataset as reported by \citet{mteb}.

\noindent \textbf{Binary Pair Classification:}
\begin{itemize}
\itemsep0em 
\item \textbf{SprintDuplicateQuestions (SDQ)} \cite{mteb_SDQ}.
Collection of questions from the Sprint community. The goal is to classify a pair of sentences as duplicates or not.
\item \textbf{TwitterSemEval2015 (TSE)} \cite{mteb_tse}.
Paraphrase-Pairs of Tweets from the SemEval
2015 workshop. The goal is to classify a pair of
tweets as paraphrases or not.
\item \textbf{TwitterURLCorpus (TUC)} \cite{mteb_tuc}.
Paraphrase-Pairs of Tweets. The goal is to
classify a pair of tweets as paraphrases or not.
\end{itemize}

\noindent \textbf{Multi-class Single-input Classification:}
\begin{itemize}
\itemsep0em 
\item \textbf{EmotionClassification (EC)} \cite{mteb_ec}. Dataset of English Twitter messages with six basic emotions: anger,
fear, joy, love, sadness, and surprise.
\item \textbf{TweetSentimentExtractionClassification (TSEC)}.\footnote{\url{https://www.kaggle.com/competitions/
tweet-sentiment-extraction}} 
TweetSentimentExtraction Dataset from Kaggle competition. Sentiment classification of
tweets as neutral, positive or negative.
\end{itemize}

\begin{table*}[ht]
\fontsize{8}{8}
\selectfont
\centering
\renewcommand{\arraystretch}{1.7} 
\tabcolsep=0.06cm
\begin{tabular}[]{ll|ll|ll|lll|ll|ll}
\hline
Task Type&&\multicolumn{6}{c}{Binary Pair Class.}
&&\multicolumn{4}{c}{Single Input Class.}
\\ \hline
Dataset&&\multicolumn{2}{c}{SDQ}&\multicolumn{2}{c}{TSE}&\multicolumn{2}{c}{TUC}&&\multicolumn{2}{c}{EC}&\multicolumn{2}{c}{TSEC} \\ \hline
Metric&& acc. & m.f1  & acc. & m.f1  & acc. & m.f1 
&& acc. & m.f1  & acc. & m.f1 \\ \hline
\multicolumn{13}{l}{\textbf{RQ1: Are fine-tuned LLMs good classifiers?}}\\ \hline
\begin{tabular}{l}\hspace{1ex}(1). The best results are achieved with the SeqC approach.\end{tabular}
&&\cellcolor{green!20}y&\cellcolor{green!20}y
&\cellcolor{green!20}y&\cellcolor{green!20}y
&\cellcolor{green!20}y&\cellcolor{green!20}y
&\cellcolor{green!20}&\cellcolor{green!20}y&\cellcolor{green!20}y
&\cellcolor{green!20}y&\cellcolor{green!20}y
\\ \hline
\begin{tabular}{l}\hspace{1ex}(2). The best results outperform fully fine-tuned PLMs.\end{tabular}
&&\cellcolor{green!20}y&\cellcolor{green!20}y
&\cellcolor{green!20}y&\cellcolor{green!20}y
&\cellcolor{green!20}y&\cellcolor{green!20}y
&\cellcolor{green!20}&\cellcolor{green!20}y&\cellcolor{green!20}y
&\cellcolor{green!20}y&\cellcolor{green!20}y
\\ \hline
\begin{tabular}{l}\hspace{1ex}(3). The best results set a new SOTA.\end{tabular}
&&\cellcolor{green!20}y&\cellcolor{green!20}y
&\cellcolor{green!20}y&\cellcolor{green!20}y
&\cellcolor{green!20}y&\cellcolor{green!20}y
&\cellcolor{green!20}&\cellcolor{green!20}y&\cellcolor{yellow!20}n
&\cellcolor{yellow!20}n&\cellcolor{yellow!20}n
\\ \hline

\begin{tabular}{l}*(4). Using SeqC, LLMs outperform a fully fine-tuned RoBERTa.\end{tabular}
&&\cellcolor{green!50}${\frac 7 8}$&\cellcolor{green!50}${\frac 7 8}$
&\cellcolor{green!50}${\frac 7 8}$&\cellcolor{green!50}${\frac 7 8}$
&\cellcolor{green!50}${\frac 8 8}$&\cellcolor{green!50}${\frac 8 8}$
&\cellcolor{green!50}
&\cellcolor{yellow!20}${\frac 5 8}$&\cellcolor{red!15}${\frac 2 8}$
&\cellcolor{red!15}${\frac 3 8}$&\cellcolor{yellow!20}${\frac 4 8}$
\\ \hline
\multicolumn{13}{l}{\textbf{RQ2: Which LLMs are more effective as classifiers?}}\\ \hline
\begin{tabular}{l}\hspace{1ex}(1). The 13B llama2 models or the 8B llama3 models produce \\the best results. \end{tabular}
&&\cellcolor{green!20}y&\cellcolor{green!20}y
&\cellcolor{green!20}y&\cellcolor{green!20}y
&\cellcolor{green!20}y&\cellcolor{green!20}y
&\cellcolor{green!20}&\cellcolor{green!20}y&\cellcolor{green!20}y
&\cellcolor{green!20}y&\cellcolor{green!20}y
\\ \hline
\begin{tabular}{l}*(2). Using Gen, instruction-fine-tuned LLMs outperform their \\non-instruction-fine-tuned counterparts.\end{tabular}
&&\multicolumn{6}{c}{\cellcolor{red!15}n}
&\cellcolor{red!15}&\multicolumn{4}{c}{\cellcolor{green!50}y}
\\ \hline

\multicolumn{13}{l}{\textbf{RQ3: Which approach is most effective?}}\\ \hline

\begin{tabular}{l}*(1). In terms of performance, SeqC (and XNet) is most effective. \end{tabular}
 &\\ 
\begin{tabular}{l}\quad a. SeqC outperforms Gen  \end{tabular}
&&\cellcolor{green!50}${\frac 7 8}$&\cellcolor{green!50}${\frac 7 8}$
&\cellcolor{green!20}${\frac 7 8}$&\cellcolor{green!20}${\frac 7 8}$
&\cellcolor{green!50}${\frac 7 8}$&\cellcolor{green!50}${\frac 7 8}$
&\cellcolor{green!50}
&\cellcolor{green!50}${\frac 8 8}$&\cellcolor{green!50}${\frac 8 8}$
&\cellcolor{green!50}${\frac 7 8}$&\cellcolor{green!50}${\frac 7 8}$
\\ 
\begin{tabular}{l}\quad b. SeqC outperforms SNet\end{tabular}
&&\cellcolor{green!50}${\frac 7 8}$&\cellcolor{green!50}${\frac 7 8}$
&\cellcolor{green!50}${\frac 8 8}$&\cellcolor{green!20}${\frac 7 8}$
&\cellcolor{green!50}${\frac 8 8}$&\cellcolor{green!50}${\frac 8 8}$
&\cellcolor{green!50}&-&-&-&-\\ 
\begin{tabular}{l}\quad c. XNet outperforms SNet \end{tabular}
&&\cellcolor{green!50}${\frac 7 8}$&\cellcolor{green!50}${\frac 7 8}$
&\cellcolor{green!50}${\frac 7 8}$&\cellcolor{green!50}${\frac 7 8}$
&\cellcolor{green!50}${\frac 8 8}$&\cellcolor{green!50}${\frac 8 8}$
&\cellcolor{green!50}
&-&-
&-&-
\\ 
\begin{tabular}{l}\quad d. In terms of performance, there are  no significant differences \\\quad between SeqC and XNet.\end{tabular}
&&\cellcolor{green!50}y&\cellcolor{green!50}y
&\cellcolor{green!50}y&\cellcolor{green!50}y
&\cellcolor{green!50}y&\cellcolor{green!50}y
&\cellcolor{green!50}
&-&-
&-&-
\\ \hline
Metric&& eff. & AIR  & eff. & AIR  & eff. & AIR 
&& eff. & AIR  & eff. & AIR \\ \hline
\begin{tabular}{l}*(2). In terms of inference efficiency (eff.), SeqC is most effective.\end{tabular}
 &\\ 
\begin{tabular}{l}\quad a. SeqC > Gen  \end{tabular}
&&\cellcolor{green!50}${\frac 8 8}$&\cellcolor{green!50}
&\cellcolor{green!50}${\frac 8 8}$&\cellcolor{green!50}
&\cellcolor{green!50}${\frac 8 8}$&\cellcolor{green!50}
&\cellcolor{green!50}
&\cellcolor{green!50}${\frac 8 8}$&\cellcolor{green!50}
&\cellcolor{green!50}${\frac 8 8}$&\cellcolor{green!50}
\\ 
\begin{tabular}{l}\quad b. SeqC > SNet \end{tabular}
&&\cellcolor{green!50}${\frac 8 8}$&\cellcolor{green!50}
&\cellcolor{green!50}${\frac 8 8}$&\cellcolor{green!50}
&\cellcolor{green!50}${\frac 8 8}$&\cellcolor{green!50}
&\cellcolor{green!50}
&-&
&-&
\\
\begin{tabular}{l}\quad c. SeqC > XNet  \end{tabular}
&&\cellcolor{green!50}${\frac 7 8}$&\cellcolor{green!50}
&\cellcolor{green!50}${\frac 8 8}$&\cellcolor{green!50}
&\cellcolor{green!50}${\frac 6 8}$&\cellcolor{green!50}
&\cellcolor{green!50}
&-&
&-&\\
\begin{tabular}{l}\quad d. XNet > SNet \end{tabular}
&&\cellcolor{green!50}${\frac 8 8}$&\cellcolor{green!50}
&\cellcolor{green!50}${\frac 8 8}$&\cellcolor{green!50}
&\cellcolor{green!50}${\frac 8 8}$&\cellcolor{green!50}
&\cellcolor{green!50}
&-&
&-&
\\ \hline
\begin{tabular}{l}\hspace{1ex}(3). Generative models encounter AIR issues even after fine-tuning. \end{tabular}
&&\cellcolor{red!15}&\cellcolor{red!15}${\frac 0 8}$
&\cellcolor{green!20}&\cellcolor{green!20}${\frac 1 8}$
&\cellcolor{green!20}&\cellcolor{green!20}${\frac 1 8}$
&\cellcolor{green!20}&\cellcolor{green!20}&\cellcolor{green!20}${\frac 6 8}$
&\cellcolor{green!20}&\cellcolor{green!20}${\frac 3 8}$
\\ \hline
\end{tabular}
\caption[]{Summary of research questions, findings, and results of significance tests. A "y" indicates that the finding generalizes to the respective task, while the values represent the count of LLMs among the eight tested that support the statement. Statements marked with an asterisk (*) indicate that significance tests are applicable, and the dark green color denotes significant support for the statistical tests with p<0.05. Statements regarding model performance are evaluated based on accuracy (acc.) and macro average F1 scores (m. f1), respectively. The yellow color indicates that the performance is close to SOTA, or that the statement is supported by >= 50\% of the LLMs. Further details are provided in §\ref{subsec:gen_res}}
\label{tab:gen_test}
\end{table*}

\subsection{Experimental Details}
\label{subsec:gen_exp}
The binary pair classification datasets are imbalanced, containing a substantial number of negative samples. To balance our training data, we retain positive samples from the original training set and randomly select an equivalent number of negative samples, resulting in 1,786 training samples for the SDQ dataset, 5,494 for TSE and 5,000 for the TUC dataset. For each dataset, we randomly select 1k validation samples and 2k test samples, ensuring that the original imbalanced label proportions are maintained.
For the single-input classification datasets, we randomly select 16k training samples, along with 1k validation samples and 2k test samples, all preserving the original label proportions. 

We fine-tune all linear layers of the LLMs using QLoRA \cite{qlora}, maintaining the same LoRA rank ($r$), LoRA alpha ($a$), and dropout ($d$) parameters as we used for the EIC task (see §\ref{sec:exp_details}). However, with the smaller SDQ dataset, the results are consistently suboptimal; we thus adjust the parameters for the SDQ dataset as follows: for approach \textit{Gen}, we set $r$=64, $a$=64, $d$=0.1; for approaches \textit{SeqC}, \textit{SNet}, and \textit{XNet}, the settings are $r$=32, $a$=32, $d$=0.1. 

On the binary pair classification datasets, the models are fine-tuned for ten epochs on the training set, with checkpoints saved after each epoch. For the larger EC and TSEC datasets, we fine-tune the models for five epochs. The final model checkpoint is selected based on evaluation results from the validation set, and the model's performance is subsequently assessed on the test set. The computational resources utilized are the same as those employed for the EIC task, as reported in §\ref{sec:exp_details}.

\subsection{Results and Discussion}
\label{subsec:gen_res}
Table \ref{tab:gen}  presents the results for the five additional tasks using the four approaches with the eight LLMs and two PLMs.\footnote{For each approach, we utilize the best input and transformation function settings observed from the EIC task for LLMs and PLMs fine-tuning. For \textit{Gen}, we employ \textcircled{\tiny 2} \textit{inst + structured input}. For \textit{SeqC}, \textit{XNet}, and \textit{SNet}, structured inputs are used for pair classification datasets, while simple natural inputs are used for single-input datasets. The transformation function \textit{diffABS} is applied for both \textit{XNet} and \textit{SNet}.} Previous SOTA results, as of Sept. 2024, are reported according to the MTEB benchmark leaderboard\footnote{\url{https://huggingface.co/spaces/mteb/leaderboard}} and the SOTA model evaluation results.\footnote{\url{https://huggingface.co/nvidia/NV-Embed-v2}}
Based on these results, we address the following three research questions to determine whether the observations from the EIC task generalize to the new tasks. Table \ref{tab:gen_test} summarizes the research questions, findings, and results of the significance tests.

\noindent \textbf{RQ1: Are fine-tuned LLMs good classifiers?} 
It is confirmed in the additional tasks that LLMs can be effectively enhanced to serve as good classifiers, achieving results that are better than or comparable to those of previous SOTA models and fully fine-tuned PLMs. 
First, we observe that across all datasets, the best results  (highlighted in bold in Table \ref{tab:gen}) are achieved by fine-tuning LLMs using the \textit{SeqC} approach. These results outperform fully fine-tuned PLMs across all datasets and establish new SOTA performance for the pair classification tasks. In the single-input tasks, our best results are comparable to the SOTA results.
Next, we examine the fine-grained results for each LLM. Using the \textit{SeqC} approach on the pair classification datasets, most LLMs outperform the fully fine-tuned RoBERTa. One-sample t-tests indicate that the higher-performing LLMs are significantly better than the RoBERTa baseline.
These findings suggest that LLMs possess superior encoding capabilities and can be fine-tuned as good classifiers using the encoding-based \textit{SeqC} approach, with particularly substantial potential for pair classification tasks.

\noindent \textbf{RQ2: Which LLMs are more effective as classifiers?} As shown in Table \ref{tab:gen}, across all five additional tasks, the best results (highlighted in bold) are achieved by the 13B Llama2 or 8B Llama3 models, reinforcing the findings from the EIC task.
Additionally, we observe that using the \textit{Gen} approach for the multi-class single-input classification tasks, instruction-fine-tuned LLMs consistently outperform their non-instruction-fine-tuned counterparts, with statistical significance supported by one-sided Wilcoxon signed-rank tests. However, no consistent or statistically significant performance differences are observed between the chat and non-chat versions of LLMs in binary pair classification tasks, particularly when the tasks are relatively straightforward and the label categories ('yes' or 'no') are easy to interpret.
These findings suggest that instruction-fine-tuned LLMs demonstrate superiority in the \textit{Gen} approach when dealing with complex label tags and tasks, likely due to their enhanced language comprehension capabilities.

\noindent \textbf{RQ3: Which approach is most effective?} The additional experiments confirm that the \textit{SeqC} approach is superior in terms of performance, answer inclusion rate (AIR), and inference efficiency. 

In terms of performance, \textit{SeqC} and \textit{XNet} are superior, as indicated by the four sub-statements in RQ3 (1) in Table \ref{tab:gen_test}, with significance supported by one-sided Wilcoxon signed-rank tests (a-c) and two-sample t-tests (d) on most datasets. Additionally, the statistical tests conducted on the overall results of all tasks, including those from EIC, provide significant evidence supporting this finding.

In terms of inference efficiency, the \textit{SeqC} approach is significantly superior, as demonstrated in Table \ref{tab:inf_effi} and supported by one-sided Wilcoxon signed-rank tests across all datasets (see RQ3 (2) in Table \ref{tab:gen_test}). Additionally, the statistical tests on the overall results of all tasks, including those from EIC, confirm the significance of this superiority.

In four of the five datasets, we observe that LLMs continue to face AIR issues even after fine-tuning with the \textit{Gen} approach. This issue is particularly pronounced in datasets with complex label tags.
These findings suggest that the generation-based approach is not optimal in practice due to its lack of robustness, difficulty in control, and inefficiency during inference. In contrast, the proposed encoding-based approaches, particularly \textit{SeqC}, demonstrate superiority not only in performance but also in AIR and efficiency, making them well-suited for large-scale applications.

\textbf{In conclusion, the generalization evaluations conducted across the five tasks provide compelling evidence that:} (1) LLMs can be fine-tuned to operate as good classifiers, achieving SOTA results; (2) among the eight tested LLMs, the 13B Llama2 and 8B Llama3 models exhibit the greatest potential; and (3) the encoding-based \textit{SeqC} approach proves to be the most effective, demonstrating significantly superior performance, inference efficiency, and perfect AIR.

\begin{table}[t]
\fontsize{9}{9}
\selectfont
\centering
\renewcommand{\arraystretch}{1.7} 
\tabcolsep=0.06cm
\begin{tabular}[]{l|l|l|ll|l|l}
\hline
Task Type&\multicolumn{3}{c}{Binary Pair Class.}
&&\multicolumn{2}{c}{Single Input Class.}
\\ \hline
Approach/Dataset&SDQ&TSE&TUC&&EC&TSEC \\ \hline
Gen&4.5&4.2&3.3&&3.5&3.3\\ \hline
SeqC&\textbf{19.5}&\textbf{24.6}&\textbf{20.6}&&\textbf{7.7}&\textbf{8.8} \\ \hline
SNet&3.7&4.4&4.6&&-&- \\ \hline
XNet&7.9&8.9&7.3&&-&- \\ \hline
\end{tabular}
\caption[]{Inference efficiency comparison across approaches.   Inference efficiency is measured by the number of samples processed per second. Reported values are the average efficiency of the eight tested LLMs.}
\label{tab:inf_effi}
\end{table}

\begin{table*}[ht]
\fontsize{9}{9}
\selectfont
\renewcommand{\arraystretch}{1.2} 
\tabcolsep=0.15cm
\begin{tabular}{l|ll|lll|lll|lll|lll|lll}
\hline
class &\multicolumn{2}{c}{Total} &\multicolumn{3}{c}{Grammar}
&\multicolumn{3}{c}{Clarity}&\multicolumn{3}{c}{Fact/Evidence}
&\multicolumn{3}{c}{Claim}&\multicolumn{3}{c}{Other}\\ \hline
count &\multicolumn{2}{c}{348}&\multicolumn{3}{c}{17}
&\multicolumn{3}{c}{61}&\multicolumn{3}{c}{158}
&\multicolumn{3}{c}{88}&\multicolumn{3}{c}{24}\\ \hline
metrics & Acc. & M. F1  & P & R & F1  & P & R & F1
& P & R & F1
 & P & R & F1 & P & R & F1\\ \hline
&90.5&86.4&73.9&100&85&84.1&95.1&89.2&92.6&94.9&93.8&97.4&85.2&90.9&88.2&62.5&73.2\\\hline
\end{tabular}
\caption[joint_intent]{Human evaluation of the annotated \textit{Re3-Sci2.0} dataset. Displayed are the overall accuracy (Acc.), macro average F1 score (M. F1), and precision (P), recall (R), and F1 score for each label. The failures are particularly associated with the low-resource "Other" class in the training set \cite{re3}, while the other classes have substantial F1 scores.}
\label{tab:human_eval}
\end{table*}

\section{Auto-annotation}
\label{sec:app_auto_anno}
\subsection{Revision Alignment}
\label{subsec:app_rev_align}
Both source datasets, F1000RD and NLPeer contain structured documents organized into sections and paragraphs, which we refine to sentences using the method proposed by \citet{re3}.
To manage the extensive comparison scope resulting from candidate pairs within long document revisions, we employ a two-stage approach for revision alignment.
Initially, we utilize the lightweight pre-alignment algorithm proposed by \citet{re3}, which efficiently identifies candidates and accurately extracts revision pairs with a precision of 0.99, while maintaining minimal computational cost.
However, the recall for alignment (0.92) is relatively low due to the algorithm's stringent aligning rules. To address this, we fine-tune a Llama2-13B model using approach \textit{SeqC} with instruction and structured input on the revision alignment data from Re3-Sci. This achieves a precision of 0.99 for non-alignment and a recall of 0.99 for alignment, perfectly enhancing the pre-alignment algorithm.
We selectively apply the fine-tuned model to non-aligned candidates identified by the pre-alignment algorithm. This approach allows us to identify missing revision pairs without significantly increasing computational overhead.
The identified revision pairs are annotated with the action label "Modify". Sentences in the new document that do not align with any in the old document are labeled as "Add", while unmatched sentences in the old document are marked as "Delete".

\subsection{Human Evaluation}
\label{subsec:app_human_eval}
A human evaluation of the labeled \textit{Re3-Sci2.0} data is conducted by the creator of the original Re3-Sci dataset, randomly selecting 10 documents with 348 edits. 
The evaluation reveals 100\% accuracy for revision alignment, and for edit intent classification, a 90.5\% accuracy and a macro average F1 score of 86.4. Table \ref{tab:human_eval} indicates that the failures in edit intent classification are particularly associated with the low-resource "Other" class in the training set \cite{re3}, while the other classes have substantial F1 scores.

\subsection{Subject Domains and Document Categories}
\label{subsec:app_area_cat}
The F1000RD documents fall into three main subject domains according to the F1000RD website\footnote{\url{https://f1000research.com/}, as of April 2024}: 

\begin{itemize}
    \item Medical and health sciences focuses on the provision of healthcare, the prevention and treatment of human diseases and interventions and technology for use in healthcare to improve the treatment of patients.
    \item Natural sciences comprises the branches of science which aim to describe and understand the fundamental processes and phenomena that define our natural world, including both life sciences and physical sciences.
    \item Social sciences subject areas seeks to understand social relationships, societal issues and the ways in which people behave and shape our world.
\end{itemize}

The six document categories are defined as:
\begin{itemize}
    \item \textit{nlp}: documents from the NLPeer corpus that present research on Natural Language Processing 
    \item \textit{case (med)}: specific F1000RD documents from the medical and health sciences that provide short reports on individual medical cases
    \item \textit{med}: other research papers from the medical and health sciences domain within the F1000RD dataset
    \item \textit{tool (nat)}: specific  F1000RD  documents from the natural sciences domain that provide technical reports on software or tools, primarily from computational biology
    \item \textit{nat}: other research papers from the natural sciences field within the F1000RD dataset
    \item \textit{soc}: documents from the social sciences domain within the F1000RD dataset
\end{itemize}
Documents that do not fit into any domains or belong to more than one domain are excluded from the divisions.

\begin{table}[t]
\fontsize{9}{9}
\selectfont
\centering
\renewcommand{\arraystretch}{1.22} 
\begin{tabular}[]{lllllll}
\toprule 
&\textit{successful}&\textit{unsuccessful}\\\hline
\#Grammar&5.5&6.1\\
\#Clarity&\textbf{9.3}&\textbf{7.3}\\
\#Fact/Evidence&22.0&19.1\\
\#Claim&\textbf{8.6}&\textbf{5.9}\\
\#Other&1.0&0.7\\
\#edits&\textbf{46.4}&\textbf{39.1}\\
\bottomrule
\end{tabular}
\caption[]{Average number of edits per intent per document and average number of total edits per document. Values are bolded if two-sample t-tests indicate a significant difference between the successful and unsuccessful groups,  with p<0.05.}
\label{tab:frd_success_fail_count}
\end{table}

\section{Edit Analysis}
\label{sec:app_meta_analysis}

\subsection{Successful vs. Unsuccessful Revisions}
\label{subsec:app_success_fail}
We interpret increased reviewer scores as indicators of successful revisions and improvements in scientific quality. 
Reviewers in the F1000RD community evaluate publications using one of three decisions: "reject," "approve-with-reservations," or "approve", which we convert into numeric values.\footnote{"reject":1, "approve-with-reservations":2, "approve":3} Document revisions that result in an increased average reviewer score are considered successful, while those that do not are deemed unsuccessful. Among the 849 F1000RD documents with reviewer scores for both initial and final versions, 575 are categorized as successful and 274 as unsuccessful. 
Documents from the NLPeer corpus lack final reviewer scores for their final versions; however, since all are accepted to a venue, we assume that the 325 documents have all undergone successful revisions. Given that our objective for RQ1 in §\ref{sec:meta_analysis} is to compare successful revisions with unsuccessful ones, we utilize the categorized F1000RD documents for the analysis, as the NLPeer documents lack unsuccessful samples.

Table \ref{tab:frd_success_fail_count} shows that successful revisions contain significantly more edits than unsuccessful ones, particularly with more changes in Clarity and Claim.

\subsection{Editing Behavior across Research Domains and Document Categories}
\label{subsec:app_rev_area}

\begin{figure}[ht]
\centering
  \begin{subfigure}[b]{0.4\textwidth}
    \includegraphics[width=\linewidth]{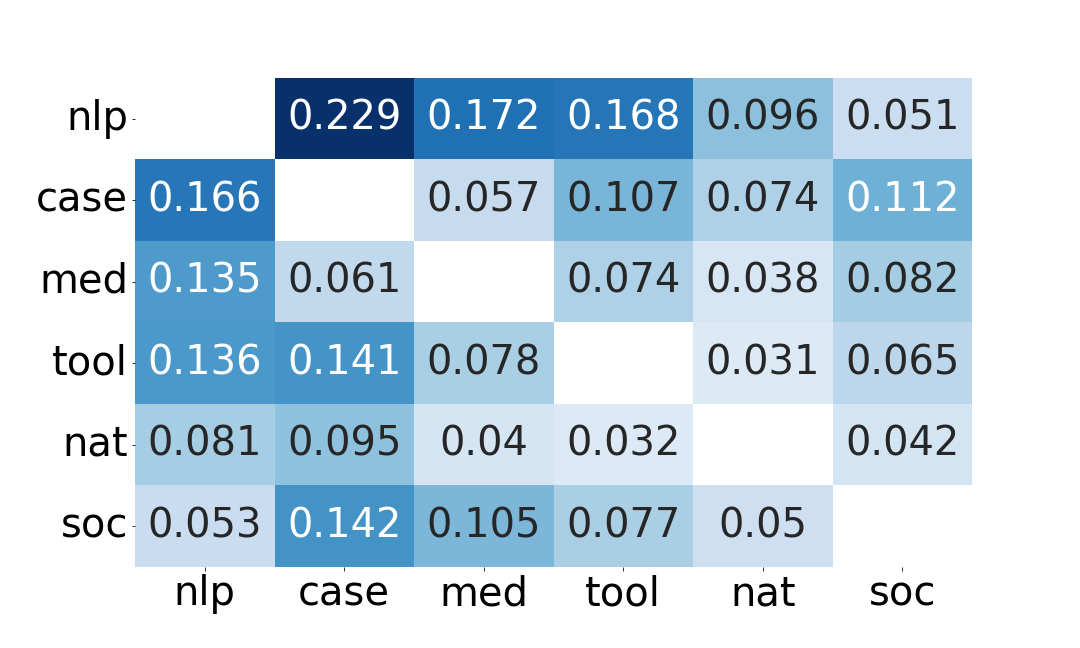}
    \caption{action location}
    \label{fig:ea_loc_kl}
  \end{subfigure}
  \begin{subfigure}[b]{0.4\textwidth}
    \includegraphics[width=\linewidth]{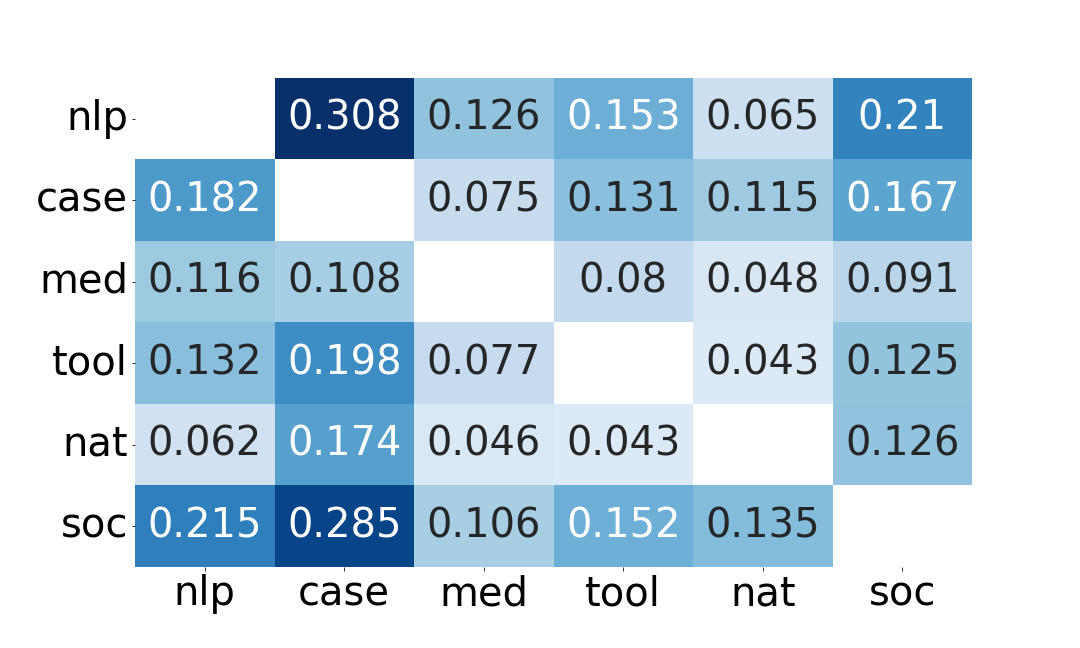}
    \caption{intent location}
    \label{fig:ei_loc_kl}
  \end{subfigure} 
  \begin{subfigure}[b]{0.4\textwidth}
    \includegraphics[width=\linewidth]{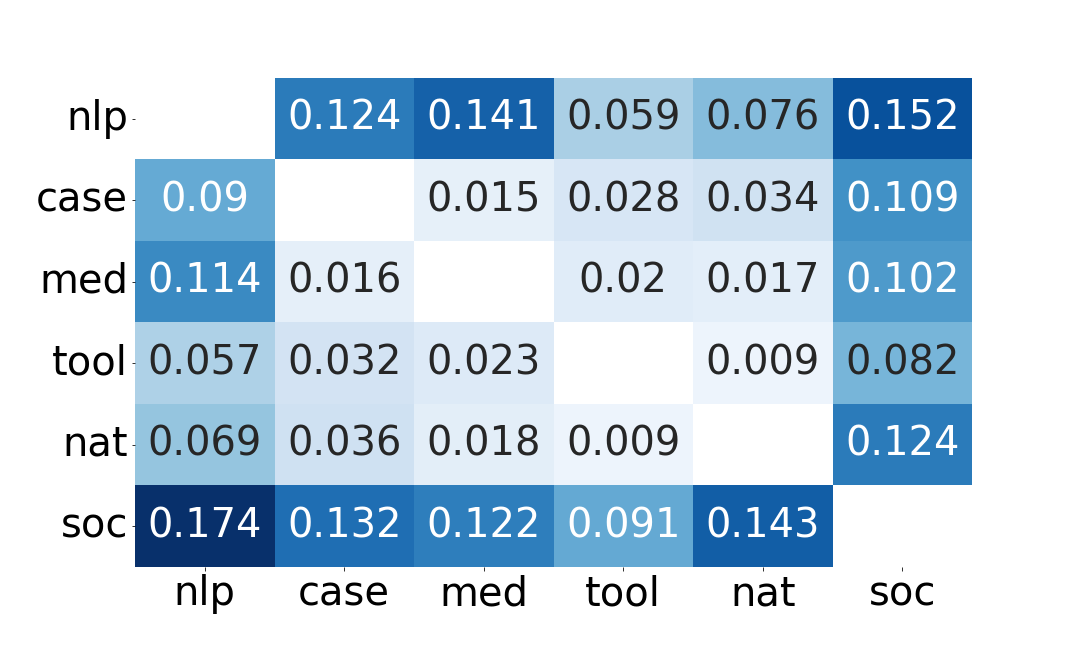}
    \caption{label combination}
    \label{fig:ea_ei_kl}
  \end{subfigure} 
  \caption{Kullback–Leibler (KL) Divergence analysis of the distributions across categories for  (a) action location (Figure \ref{fig:loc}, 1st line) (b) intent location (Figure \ref{fig:loc}, 2nd line) and (c) edit action and intent combinations (Figure \ref{fig:ea_and_ei}). The higher the KL divergence, the greater the difference between the distributions.}
  \label{fig:KL}
\end{figure}

\end{document}